\documentclass[preprint,12pt]{elsarticle}

\usepackage{amssymb}
\usepackage{amsmath}
\usepackage[ruled,vlined,linesnumbered]{algorithm2e}            
\biboptions{square,numbers,sort&compress}
\usepackage{booktabs}
\usepackage{makecell}
\usepackage{multicol}
\usepackage{multirow}
\usepackage{tabularx}
\usepackage{caption}
\usepackage{float}
\usepackage{graphicx}
\usepackage{longtable}
\usepackage{xltabular}
\usepackage{enumitem}
\usepackage{tabularx}
\usepackage{geometry}
\usepackage{rotating}
\usepackage{threeparttable}
\usepackage{array} 
\usepackage{afterpage}

\newcolumntype{C}[1]{>{\centering\arraybackslash}p{#1}}
\usepackage{booktabs} 
\newcolumntype{R}[1]{>{\raggedright\arraybackslash}p{#1}}
\usepackage{caption} 
\graphicspath{ {./Figs/} }

\journal{Decision Analytics Journal}

\begin{document}

\begin{frontmatter}



\title{Towards a General Framework for Predicting and
Explaining the Hardness of Graph-based Combinatorial
Optimization Problems using Machine Learning and
Association Rule Mining}


\author[inst1]{Bharat S. Sharman\fnmark[1]}
\affiliation[inst1]{organization={School of Computational Science and Engineering},
            addressline={McMaster University}, 
            city={Hamilton},
            postcode={L8S 4E8}, 
            state={ON},
            country={Canada}}

\author[inst2]{Elkafi Hassini}
\affiliation[inst2]{organization={DeGroote School of Business},
            addressline={McMaster University}, 
            city={Hamilton},
            postcode={L8S 4E8}, 
            state={ON},
            country={Canada}}

\fntext[1]{Corresponding author: sharmanb@mcmaster.ca (Bharat S. Sharman)}

\begin{abstract}
This study introduces GCO-HPIF, a general machine learning-based framework to predict and explain the computational hardness of combinatorial optimization problems that can be represented on graphs. The framework consists of two stages. In the first stage, a dataset is created comprising problem-agnostic graph features and hardness classifications of problem instances. Machine learning-based classification algorithms are trained to map graph features to hardness categories. In the second stage, the framework explains the predictions using an association rule mining algorithm. Additionally, machine learning-based regression models are trained to predict algorithmic computation times. The GCO-HPIF framework was applied to a dataset of 3287 maximum clique problem instances compiled from the COLLAB, IMDB and TWITTER graph datasets using five state-of-the-art algorithms namely, three exact Branch and Bound-based algorithms Gurobi, CliSAT and MOMC and two graph neural network-based algorithms EGN and HGS. The framework demonstrated excellent performance in predicting instance hardness, achieving a weighted F1-score of 0.9921, a minority class F1-score of 0.878 and an ROC-AUC score of 0.9083 using only three graph features. The best association rule found by the FP-Growth association rule mining algorithm for explaining the hardness predictions had a support of 0.8829 for hard instances and an overall accuracy of 87.64\%, underscoring the framework's usefulness for both prediction and explanation. Furthermore, the best performing regression model for predicting computation times achieved a percentage RMSE of 5.12 and an $R^2$ value of 0.991 respectively.    
\end{abstract}

\begin{graphicalabstract}
\includegraphics[width=15.5cm,height=12cm]{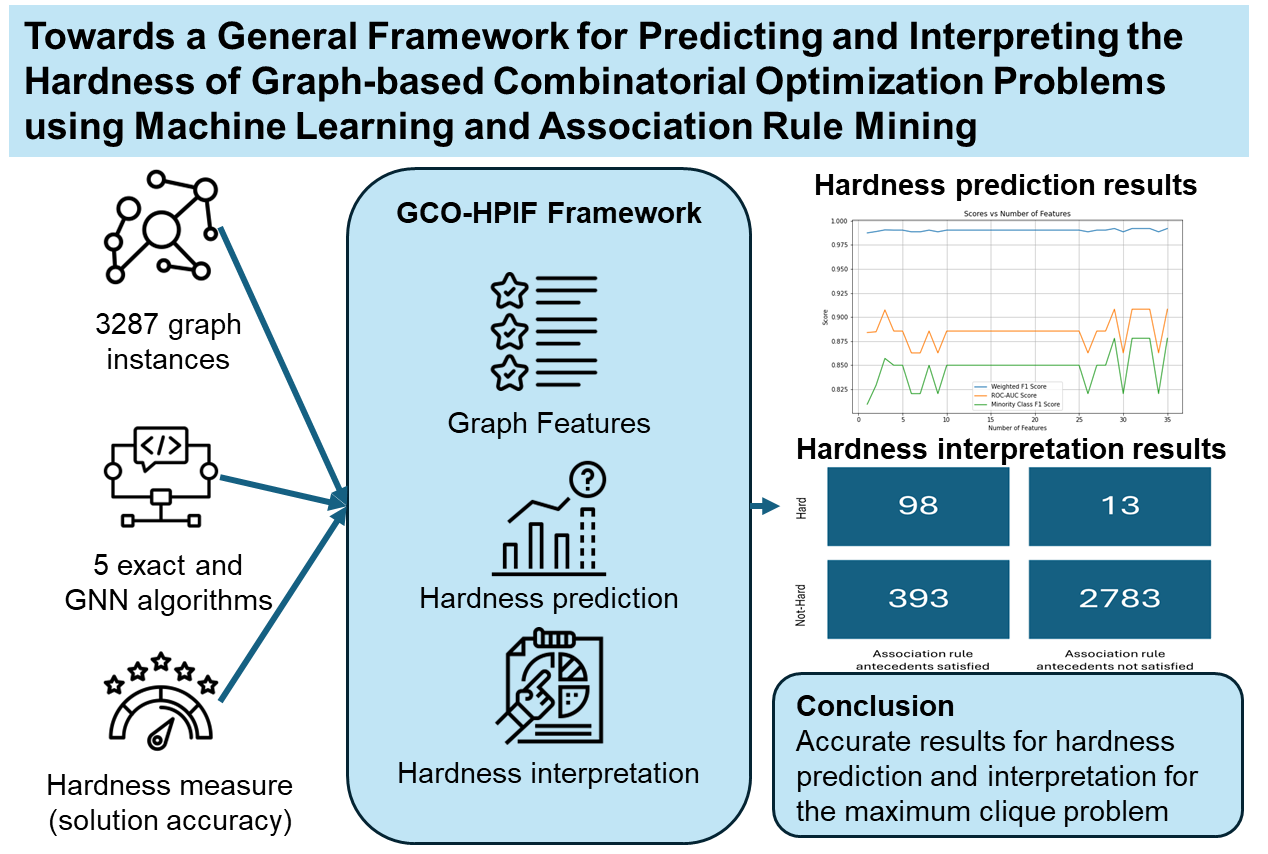}
\end{graphicalabstract}


\begin{highlights}
\item Novel framework to explain hardness of graph combinatorial optimization problems.
\item Generic, problem-agnostic polynomial time computable graph-based features used.
\item Excellent performance on predicting the hardness of maximum clique problem.
\item Interpretable rules having high support derived using association rule mining.
\end{highlights}


\end{frontmatter}



\section{Introduction}
\label{sec:intro}

Combinatorial optimization problems (COPs) involve selecting a subset of objects from a finite set to optimize a specified objective function \cite{du2022introduction}. These problems are of substantial theoretical and practical relevance. Their theoretical significance stems from the fact that many COPs, such as the traveling salesperson problem (TSP), are known to be NP-hard and have been extensively studied to explore their theoretical properties \cite{pop2024comprehensive}. Practically, COPs serve as foundational tools in various fields such as supply chain management \cite{zhuang2024improving}, logistics \cite{ren2023review}, bioinformatics \cite{calvet2023role}, semiconductor chip manufacturing \cite{li2023scheduling} and auctions \cite{mansouri2019optimal} among several others. Many COPs, such as graph coloring, maximum clique (MCP), maximum independent set, minimum vertex cover, vehicle routing (VRP), and location routing (LRP) etc., can be effectively modelled using a graph data structure. This graph-based representation facilitates the use of both general-purpose algorithms, such as graph neural networks (GNN) as well as problem-specific algorithms (such as CliSAT \cite{san2023clisat} for MCP).\\

The No-Free Lunch (NFL) theorem \cite{wolpert1997no} asserts that no single algorithm can optimally solve all instances of a given problem. When considered alongside the NP-hard nature of several COPs, this theorem underscores the importance of accurately predicting the hardness of problem instances before attempting to solve them. Such predictions can significantly reduce computational effort and resource expenditure. Moreover, explainable predictions offer dual benefits. First, they enhance the conceptual understanding of the relationship between problem features and instance hardness. Second, they enable more effective deployment of algorithms by practitioners who may lack expertise in the mathematical foundations of the algorithms themselves. For instance, if hospital scheduling staff can determine the most suitable scheduling algorithm for patient procedures using few key and simple explainable rules derived from problem features, then they can efficiently plan operations without requiring assistance from computational scientists. \\

Several studies have focused on investigating the hardness of graph-based COPs. A review of these studies is presented in Section \ref{sec:lit_review}. The majority of these works examine specific COPs and utilize problem-specific features to characterize or predict instance hardness. While this approach can yield SOTA results for individual problems, it presents challenges for generalizing insights into problem hardness and for facilitating comparisons across different COPs. This limitation arises because features tailored to a specific problem may lack relevance or even applicability in the context of other problems. Furthermore, there has been limited research on developing explainable predictions of instance hardness. explainabiilty not only deepens the understanding of but also fosters greater user trust in the algorithms by providing transparent insights into the underlying decision-making processes. \\
This study seeks to address these research gaps by introducing a machine learning (ML)-based framework for predicting the hardness of graph-based COPs. The framework leverages generic, problem-agnostic graph features, including spectral features as inputs, and provides explainable results through the application of association rule mining (ARM). The key contributions of this study are as follows: \\

\begin{itemize}
    \item Developing a framework for hardness prediction in graph-based COPs that relies on generic graph features, enabling problem-agnostic applicability.   
    \item Developing a method of explaining hardness prediction results using ARM, enhancing transparency and understanding of the prediction process.
    \item Demonstrating the proposed framework's effectiveness by applying it to the MCP, a canonical NP-hard graph-based COP.   
\end{itemize}

This paper is organized as follows: Section \ref{sec:gb-cop} introduces graph-based COPs, providing the foundational context for the study. A review of prior research on evaluating the hardness of graph-based COPs is presented in Section \ref{sec:lit_review}. Section \ref{sec:significance_generic_graph_features} discusses the importance of utilizing generic graph features for developing a generalizable framework for predicting and explaining the hardness of such problems. The proposed framework is detailed in Section \ref{sec:framework_hardness_cop}. Section  \ref{sec:explaining_hardness_cop_association_rule_mining} focuses ARM-based explanation of hardness prediction results. The application and performance of the framework are demonstrated using the MCP as a case study in Sections \ref{sec:application_of_framework_mcp} and \ref{sec:results_and_discussion} respectively. Finally, conclusions and directions for further research are presented in Section \ref{sec:conclusion_and_further_research}.    

\section{Graph-based combinatorial optimization problems}\label{sec:gb-cop}

Given a graph \( G = (V, E) \), where \( V = \{v_1, v_2, \dots, v_n\} \) is the set of vertices and \( E \subseteq V \times V \) is the set of edges, the general graph-based COP can be formulated as:

\begin{equation}
\begin{aligned}
    \text{Optimize:} \quad & f(S) = \sum_{i \in V \cup E} c_i x_i \\
    \text{subject to:} \quad & \sum_{i \in \mathcal{C}_j} a_{ij} x_i \leq b_j, \quad \forall j \in J \\
    & x_i \in \{0, 1\}, \quad \forall i \in V \cup E
\end{aligned}
\label{eq:graph_cop}
\end{equation}

where \( S \subseteq V \cup E \) is a subset of vertices and edges, \( c_i \) is the cost (or weight) associated with element \( i \), \( x_i \) is a binary decision variable indicating whether element \( i \) (a node or edge) is included in the solution \( S \), \( J \) is the index set for the constraints (i.e., each \( j \) corresponds to a specific constraint), \( \mathcal{C}_j \subseteq V \cup E \) is the subset of graph elements that are involved in constraint \( j \), \( a_{ij} \) is a coefficient matrix, indicating how much element \( i \) contributes to constraint \( j \), and \( b_j \) is the upper bound for constraint \( j \).

Essentially, the formulation in expression \ref{eq:graph_cop} means that the problem involves selecting a subset of vertices or edges from the graph such that the objective function $f(S)$ is optimized, with the selection being subject to a set of constraints.

\begin{figure}[h]
    \centering
    \includegraphics[width=\textwidth]{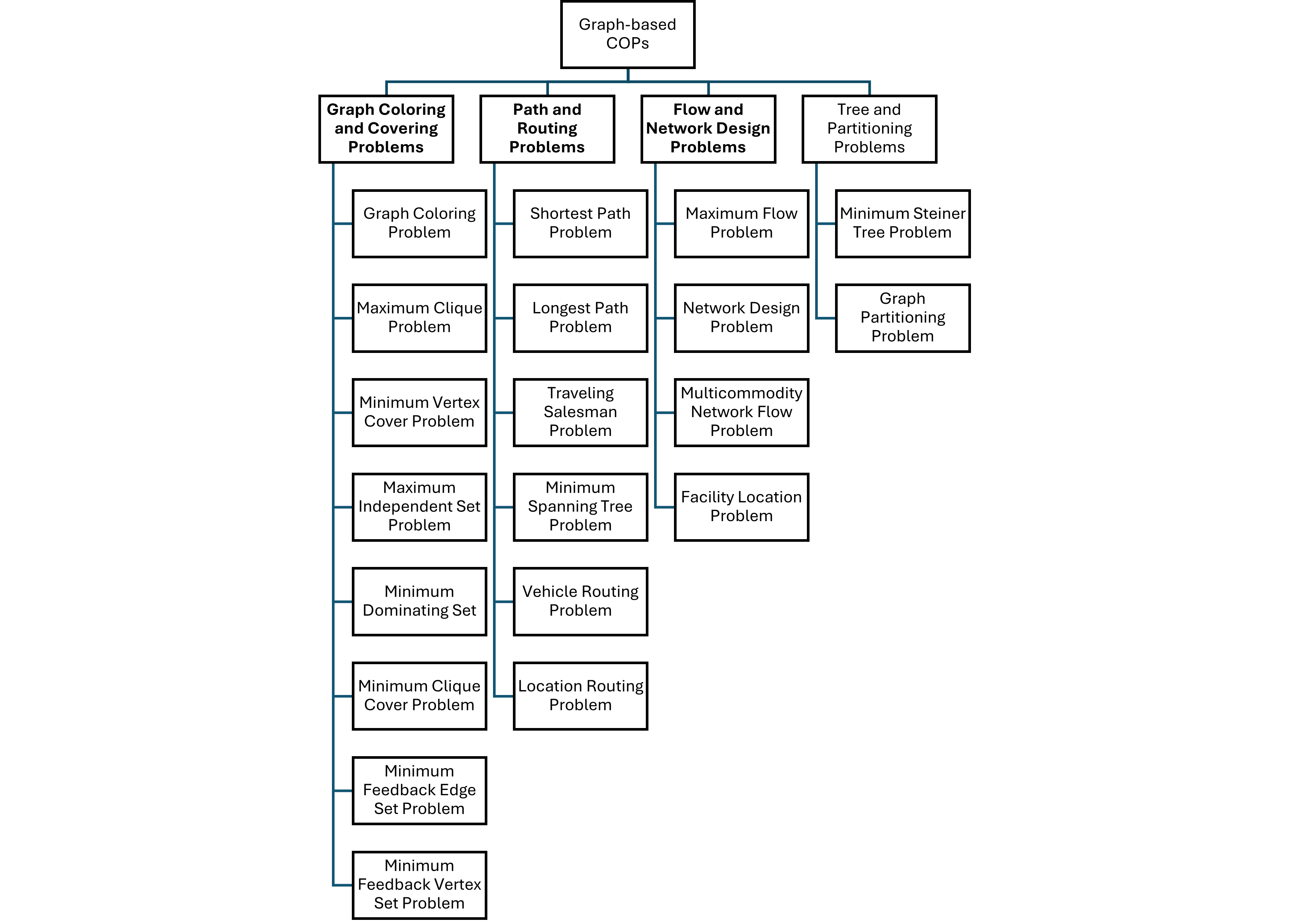}
    \caption{Categories of graph-based combinatorial optimization problems}
    \label{fig:Graph_COP_problems_classification}
\end{figure}

Numerous COPs can be modelled as graph-based problems and can therefore be solved using graph algorithms. These problems can be broadly classified into four classes, as illustrated in Figure \ref{fig:Graph_COP_problems_classification}.

 A brief description of each class is provided below:

\begin{enumerate}

    \item Graph coloring and covering problems: This category encompasses problems that involve assigning colors or labels to the vertices of a graph while satisfying specific constraints. For example, the maximum independent set problem (MISP) entails identifying the largest possible subset of nodes such that no two vertices in the subset are adjacent. 
    
    \item Path and routing problems: These problems focus on determining optimal paths or routes within a network. For instance, the location routing problem (LRP) involves selecting nodes and edges in a manner that minimizes an objective function such as cost, while meeting certain constraints.

    \item Flow and network design problems: Problems in this category are concerned with optimizing the flow of resources or information through a network. An example is the multi-commodity network flow problem (MCNFP) which involves routing multiple \lq commodities\rq (which could also include materials, energy, even people) through a network to optimize an objective function under specific constraints. While path and routing problems emphasize identifying optimal routes, flow and network design problems prioritize the efficient  allocation of resources across a network or configuring a network to achieve particular objectives.   
    
    \item Tree and partitioning problems: This class includes problems that involve partitioning a graph into sub-graphs or components. For instance, the graph partitioning problem (GPP) requires dividing a graph into partitions such that a criterion, such as minimizing the number of connections between partitions, is optimized. 
\end{enumerate}

\section{Literature review}\label{sec:lit_review}

Numerous studies have examined the hardness of COPs, with the traveling salesperson problem (TSP) receiving significant attention due to its NP-hard nature and substantial theoretical and practical importance. The TSP has served as a primary benchmark for evaluating instance hardness and algorithm performance. Additionally, problems such as SAT and Knapsack have also been widely studied in this context. A hardness measure for the TSP based on the relative performance of a test algorithm compared to a surrogate algorithm was proposed by \cite{zhang2022learning}. The surrogate algorithm's parameters are computed through several gradient descent steps from the test algorithm, and the authors argue that this measure provides a lower bound on the ground-truth optimality gap. The authors of \cite{mersmann2013novel} identified the features of TSP instances that influence the performance of the opt-2 algorithm. They utilized the approximation ratio, defined as the ratio of the tour length computed by opt-2 to the optimal length obtained via the exact solver Concorde, as a measure of problem  hardness. A linear regression model to predict the hardness of a TSP instance based on the Weibull shape parameter, calculated using a Dirichlet tessellation of the instance was developed by \cite{cardenas2018creating}. The authors further employed a genetic algorithm that again takes the shape parameter as input to generate TSP instances of varying difficulty levels. A moderate correlation between certain graph invariants derived from small edge-induced sub-graphs of TSP instances and the computational time required by the Concorde solver to solve the problem was demonstrated in \cite{cvetkovic2018complexity}. A root-exponential scaling model to explain the non-exponential scaling behaviour observed in Concorde's performance on random uniform Euclidean (RUE) instances was proposed in \cite{hoos2014empirical} . A normalization methodology for TSP instance features to mitigate the influence of instance size on measures of instance hardness and algorithm performance was proposed by the authors in \cite{heins2023study}. In contrast, \cite{huerta2022improving}, \cite{seiler2020deep} and \cite{zhao2021towards} avoided computing features explicitly. Instead, they represented TSP instances on grids and trained various convolutional neural network (CNN) models to predict solver performance and computation time, enabling algorithm selection tailored to specific TSP instances. The work of \cite{pihera2014application} introduced novel TSP instance features such as local search probing features,
geometric features, and nearest neighborhood features. They reported that these features enhanced the characterization of TSP instances and improved algorithm selection accuracy among meta-heuristic approaches. In \cite{cricsan2021randomness}, the authors posited that the spatial structure of TSP instances significantly impacts instance hardness and validated their hypothesis on two meta-heuristic algorithms, namely Ant Colony Optimization and Lin-Kernighan.\\

The SAT problem has been well-studied to analyze instance hardness and algorithm performance. \cite{lindauer2018warmstarting} leveraged insights from prior algorithm configuration experiments across various benchmark instance sets to warm-start the SparrowToRiss SAT solver. Their approach resulted in a remarkable 165-fold increase in solver speed. The use of deep exchangeable and neural
message passing deep learning algorithms for predicting the performance of SAT solver was explored in \cite{cameron2020predicting}. They demonstrated that these models consistently outperformed machine learning approaches reliant on sophisticated hand-crafted features. Researchers in \cite{chang2022predicting} highlighted the superior accuracy of graph attention networks in solving SAT problems, outperforming other neural network-based approaches.\\

Research on the Knapsack problem has focused on several aspects, including the theoretical identification of parameters that influence instance hardness \cite{jooken2022new}, the application of instance space analysis to characterize the distribution of problem instances and correlate them with algorithm performance \cite{smith2021revisiting} and the use of machine learning methods for anytime algorithm selection \cite{huerta2022improving}. These approaches rely on features known to impact instance hardness and algorithm performance.\\

Table \ref{tab:literature_review_summary} summarizes the literature reviewed and positions this study on hardness prediction and explanation within the context of existing studies. The comparison is based on four criteria, namely (a) whether the study focuses on graph-based COPs, (b)  whether the features utilized are problem-specific or problem-agnostic, (c) whether the study provides explainability for hardness predictions, and (d) whether ARM is employed to explain the hardness predictions generated by the models.

\begin{longtable}{C{5cm} C{2cm} C{2cm} C{2cm} C{2cm}}
\caption{Summary of literature review on hardness prediction and explanation for COPs} \label{tab:literature_review_summary} \\
\toprule
\thead{Study} & \thead{Graph-based \\ COP} & \thead{Problem \\ agnostic \\ features} & \thead{Hardness \\ prediction \\ explanation} & \thead{Association \\ rule \\ mining} \\
\midrule
\endfirsthead

\multicolumn{5}{c}{{\bfseries \tablename\ \thetable{} -- continued from previous page}} \\
\toprule
\thead{Study} & \thead{Graph-based \\ COP}  & \thead{Problem \\ agnostic \\ features} & \thead{Hardness \\ prediction \\ explanation} & \thead{Association \\ rule \\ mining} \\
\midrule
\endhead

\midrule \multicolumn{5}{r}{{Continued on next page}} \\
\endfoot

\bottomrule
\endlastfoot

\cite{cameron2020predicting} & &  \checkmark & &   \\
\cite{cardenas2018creating} & \checkmark & & \checkmark &   \\
\cite{chang2022predicting} & \checkmark &  & &  \\
\cite{cricsan2021randomness} & \checkmark & & \checkmark &   \\
\cite{cvetkovic2018complexity} & \checkmark &  & \checkmark &   \\
\cite{heins2023study} & \checkmark & & &    \\
\cite{hoos2014empirical} & \checkmark & & \checkmark &   \\
\cite{huerta2022improving} & \checkmark & & &   \\
\cite{jooken2022new} & & & \checkmark &    \\
\cite{mersmann2013novel} & \checkmark & & \checkmark &   \\
\cite{pihera2014application} & \checkmark & & &  \\
\cite{seiler2020deep} & \checkmark & & &    \\
\cite{smith2014exploring} & \checkmark &  \checkmark & \checkmark &   \\
\cite{smith2021revisiting} & & & \checkmark &   \\
\cite{zhao2021towards} & \checkmark & & &    \\
\cite{zhang2022learning} & \checkmark & & &   \\
Presented work & \checkmark & \checkmark & \checkmark & \checkmark \\

\end{longtable}

\section{Significance of generic graph-based features}
\label{sec:significance_generic_graph_features}

An area that has been relatively unexplored in the literature is the use of generic graph-based features for predicting the hardness of problem instances in graph-based COPs. Among the early studies in this domain is the work of \cite{smith2014exploring} which examined the influence of spectral properties on the hardness of the graph coloring problem. It has been demonstrated that numerous spectral features of graphs can be computed efficiently in polynomial time \cite{smith2012measuring}. These spectral features, being broadly applicable across various classes of graph-based COPs, offer a significant advantage over features tailored to specific problems. Leveraging such generic features, including spectral characteristics, in ML-based algorithms for hardness prediction has the potential to provide valuable insights across a wide range of graph-based COPs. This study adopts this approach in developing a framework for predicting the hardness of graph-based COPs, demonstrating its utility and generalizability. \\

Table \ref{tab:graph_features_description} enumerates the 23 graph-based and graph-level features (as against node-level features that are also graph-based but are computed for each node instead of the graph as a whole) analyzed in this study, each of which can be calculated in polynomial time. While this is not an exhaustive list of generic graph features, it still captures several important spectral and network features. The number of features were restricted to fewer than 25 in this study to avoid excessive computational time required for running the FP-Growth association rule mining that used the percentile ranges of these features as its inputs. However, this is a computational scaling challenge and more features can be added and analyzed if more compute power is available.   

\begin{longtable}{R{12cm} R{3cm}}
\caption{Graph-based features used for hardness prediction and interpretation and their computational complexity}%
\label{tab:graph_features_description} \\
\hline
\textbf{Feature Name} & \textbf{Computational Complexity} \\
\hline
\endfirsthead

\multicolumn{2}{c}{{\bfseries \tablename\ \thetable{} -- continued from previous page}} \\
\hline
\textbf{Feature Name} & \textbf{Computational Complexity} \\
\hline
\endhead

\hline \multicolumn{2}{r}{{Continued on next page}} \\ \hline
\endfoot

\hline
\endlastfoot

Number of nodes & \(O(|V|)\) \\
Number of edges & \(O(|E|)\) \\
Density & \(O(1)\) \\
Radius & \(O(|V| \cdot |E|)\) \\
Diameter & \(O(|V| \cdot |E|)\) \\
Median of the degree centralities of all nodes & \(O(|E| + |V| \log |V|)\) \\
Median of the betweenness centralities of all nodes & \(O(|V| \cdot |E|)\) \\
Median of the closeness centralities of all nodes & \(O(|V| \cdot |E|)\) \\
Global Clustering Coefficient & \(O(|V| + |E|)\) \\
Median of the eccentricities of all nodes & \(O(|V| \cdot |E|)\) \\
Algebraic Connectivity & \(O(|V|^3)\) \\
Median of the median degrees of the neighbors of a node & \(O(|V| + |E|)\) \\
Spectral Radius & \(O(|V|^3)\) \\
Laplacian Spectral Radius & \(O(|V|^3)\) \\
Median of the geodesic distance of all node pairs & \(O(|V| \cdot |E|)\) \\
Smallest non-zero eigenvalue of the Laplacian matrix & \(O(|V|^3)\) \\
Second smallest non-zero eigenvalue of the Laplacian matrix & \(O(|V|^3)\) \\
Second largest eigenvalue of the Laplacian matrix & \(O(|V|^3)\) \\
Smallest non-zero eigenvalue of the adjacency matrix & \(O(|V|^3)\) \\
Second smallest eigenvalue of the adjacency matrix & \(O(|V|^3)\) \\
Second largest eigenvalue of the adjacency matrix & \(O(|V|^3)\) \\
Difference between the largest and the second largest eigenvalues of the adjacency matrix & \(O(|V|^3)\) \\
Difference between the largest and the smallest eigenvalues of the Laplacian matrix & \(O(|V|^3)\) \\

\end{longtable}

\section{GCO-HPIF : Framework for predicting and explaining hardness of graph-based COPs}
\label{sec:framework_hardness_cop}

\begin{figure}[H]
    \centering
    \includegraphics[width=0.75\textwidth]{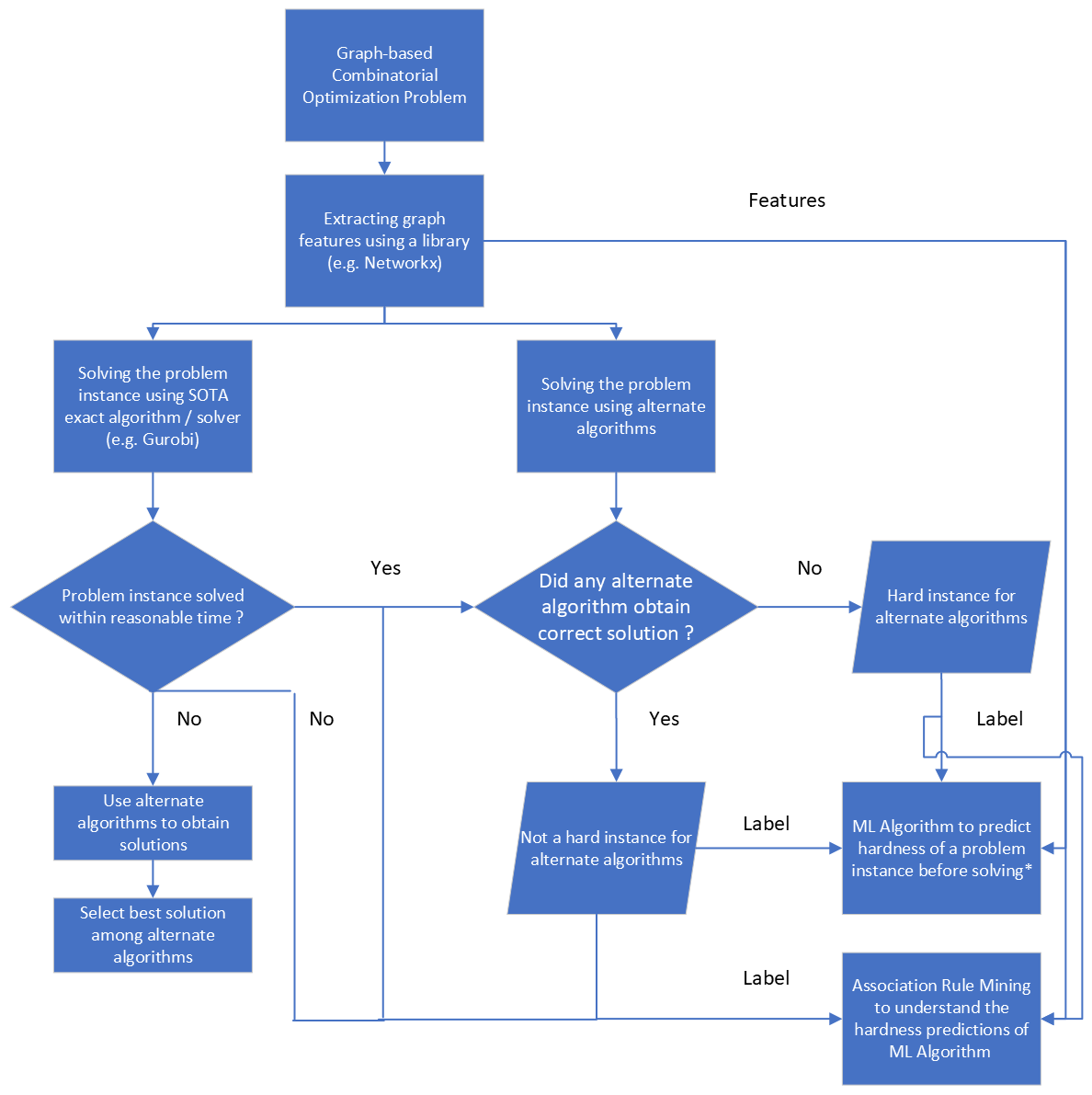}
    \caption{GCO-HPIF - a framework for predicting and explaining the hardness of graph-based COPs. *Please refer to Figure \ref{fig:ml_model_development_instance_hardness_prediction} that shows the machine learning pipeline for predicting the hardness of problem instances}
    \label{fig:Framework_for_predicting_hardness_graph_based_cop}
\end{figure}

Figure \ref{fig:Framework_for_predicting_hardness_graph_based_cop} presents the GCO-HPIF framework for predicting and explaining the hardness of graph-based COPs, where GCO-HPIF stands for \textbf{G}raph-based \textbf{C}ombinatorial \textbf{O}ptimization Problems - \textbf{H}ardness \textbf{P}rediction and \textbf{I}nterpretation \textbf{F}ramework. Initially, it is essential to verify if the COP can be directly represented as a graph. COPs like the TSP, MCP, and MISP are naturally graph-based since they involve finding paths or selecting nodes or edges. Conversely, problems such as Bin-packing and Knapsack require complex transformations to be represented graphically, which can increase computational complexity. For graph-based COPs, features computable in polynomial time, as listed in Table \ref{tab:graph_features_description}, are extracted via Networkx. The instance is solved using a SOTA exact solver such as Gurobi, alongside problem-specific algorithms. If the SOTA solver finds a solution promptly, alternative algorithms' results are compared to it. If alternatives fail to match the SOTA solution, the instance is labeled hard; if they succeed, it's labeled not hard. These labels and features train a classification ML algorithm—like XGBoost, Random Forest, or SVM—to predict instance hardness. Additionally, association rules help explain ML predictions, making the results accessible for decision-making by non-specialists. If the SOTA solver cannot solve the instance within a reasonable time limit, the best solution from the alternatives is chosen.\\

The proposed framework offers the following advantages:
\begin{enumerate}
    \item It effectively predicts the hardness of a problem instance for alternate algorithms with high accuracy prior to solving it, enhancing result accuracy when numerous instances must be processed by alternate algorithms instead of the SOTA solver for time efficiency. If the instance is deemed difficult for alternate solvers, the SOTA solver can be engaged directly, bypassing the need for alternate solvers.
    \item It aids non-experts in computational fields to easily comprehend instance hardness predictions via clear association rules.
\end{enumerate}

\begin{figure}[H]
    \centering
    \includegraphics[width=\textwidth]{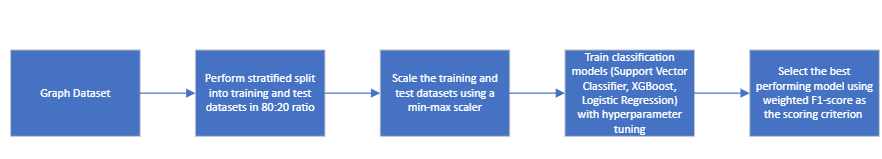}
    \caption{Steps of the machine learning model pipeline for instance hardness prediction }    \label{fig:ml_model_development_instance_hardness_prediction}
\end{figure}

\section{Explaining prediction results using association rule mining}
\label{sec:explaining_hardness_cop_association_rule_mining}

The explanation of machine learning algorithms is an actively researched topic. These algorithms are utilized for predictions across diverse areas. In critical domains like emergency care and aviation, incorrect predictions can have instant and severe repercussions. Conversely, in sectors such as job recruitment and public policy, accurate model predictions can still lead to litigation if individuals perceive discrimination from automated decision-making processes. Consequently, explainability in machine learning (xAI) has been investigated across multiple fields, including healthcare \cite{khanna2023distinctive}, \cite{alsinglawi2022explainable}, \cite{petch2022opening}, \cite{hu2022interpretable}, educational assessment \cite{hasib2022machine}, public policy \cite{bell2022s}, earthquake resistance \cite{mangalathu2022machine}, electrical infrastructure \cite{xu2022review}, credit scoring \cite{chen2024interpretable}, criminology \cite{zhang2022interpretable}, and cybersecurity \cite{nadeem2023sok}. These efforts aim to bolster decision-makers' confidence in the reliability and transparency of ML-driven models.

\subsection{Association rule mining}

Association rule mining (ARM) is a data mining method for uncovering significant relationships and patterns among items within large datasets. It is notably used in market basket analysis (MBA) for identifying co-purchase patterns \cite{unvan2021market}. Although originally developed for MBA, ARM has been applied to algorithmic prediction explanation and pattern detection in various fields, including production processes \cite{pohlmeyer2022interpretable}, power usage forecasting \cite{troncoso2023new}, runtime anomaly detection \cite{bohmer2020mining}, GPU workload monitoring \cite{li2024interpretable}, and predicting natural disasters \cite{kusak2021apriori}. ARM also aids in explaining predictions of ML algorithms like CNN-based time-series classification \cite{veerappa2023explaining}, ANN-based classification \cite{yedjour2018symbolic}, random forest-based thyroid malignancy prediction \cite{sankar2024study}, and modeling reinforcement learning behavior \cite{parham2023explaining}. Nevertheless, to the best of authors' knowledge, ARM has not yet been explored for explaining predictions of ML models on graph-based combinatorial optimization problems (COPs). This paper explores ARM's use in explainability for graph-based COPs, using MCP as a case study.

An association rule is generally expressed as $X \rightarrow Y$, where $X$ (the antecedent) is a set of items, and $Y$ (the consequent) is another set, indicating that the occurrence of $X$ suggests a likely occurrence of $Y$. In ARM, key metrics to assess the rule \( X \rightarrow Y \) are:

\begin{itemize}
    \item \textbf{Support}: The support of a rule measures the frequency with which the items in both the antecedent \( X \) and the consequent \( Y \) occur together in the dataset. It is defined as:
    \[
    \text{Support}(X \rightarrow Y) = \frac{\text{Number of transactions containing both } X \text{ and } Y}{\text{Total number of transactions}}
    \]
    Alternatively, this can be written as:
    \begin{equation}
    \text{Support}(X \rightarrow Y) = P(X \cap Y)
    \end{equation}

    \item \textbf{Confidence}: The confidence of a rule measures how often the items in the consequent \( Y \) appear in transactions that contain the antecedent \( X \). It is defined as:
    \[
    \text{Confidence}(X \rightarrow Y) = \frac{\text{Number of transactions containing both } X \text{ and } Y}{\text{Number of transactions containing } X}
    \]
    Alternatively, this can be written as:
    \begin{equation}
    \text{Confidence}(X \rightarrow Y) = P(Y \mid X) = \frac{\text{Support}(X \rightarrow Y)}{\text{Support}(X)}
    \end{equation}

    \item \textbf{Lift}: The lift of a rule measures how much more likely the items in the consequent \( Y \) are to appear in transactions that contain the antecedent \( X \) compared to their general appearance in the dataset. It is defined as:
    \[
    \text{Lift}(X \rightarrow Y) = \frac{\text{Confidence}(X \rightarrow Y)}{\text{Support}(Y)}
    \]
    Alternatively, this can be written as:
    \begin{equation}
    \text{Lift}(X \rightarrow Y) = \frac{P(Y \mid X)}{P(Y)} = \frac{\text{Support}(X \rightarrow Y)}{\text{Support}(X) \times \text{Support}(Y)}
    \end{equation}
\end{itemize}

Various ARM algorithms have been developed, with some notable ones reviewed here briefly. The AIS algorithm by \cite{agrawal1993mining} stands as one of the earliest ARM methods, scanning databases multiple times to dynamically create candidate itemsets and calculate their support incrementally. Apriori, introduced by \cite{srikant1996fast}, is another significant ARM algorithm employing a bottom-up approach to iteratively generate and prune candidate itemsets based on a minimum support threshold. Its process relies on the characteristic that any subset of a frequent itemset must be frequent. The FP-Growth algorithm developed by \cite{han2000mining} addresses Apriori's limitations by using an FP-tree, which compresses dataset information into a prefix-tree structure to recursively mine frequent patterns without explicitly generating candidate itemsets. ECLAT \cite{zaki1997new} uses a vertical data format associating each item with transaction ID lists, generating frequent itemsets by intersecting these lists through depth-first search. Lastly, MLARM \cite{fortin1996object} enhances traditional ARM by discovering rules at multiple abstraction levels, such as "animal" versus "dog," using hierarchical itemsets. This study used FP-Growth due to its compact data structure, reducing memory usage, which is advantageous for large datasets.   

\subsection{FP-Growth algorithm}

The FP-Growth (Frequent Pattern Growth) algorithm efficiently extracts frequent itemsets from extensive datasets, a pivotal step in association rule mining (ARM). In contrast to the Apriori algorithm, which requires generating candidate itemsets and scanning the dataset multiple times, FP-Growth utilizes a compact data structure known as the FP-Tree (Frequent Pattern Tree) and bypasses generating candidates. This method requires fewer database scans, enhancing performance significantly, especially in dense datasets with numerous frequent patterns. The FP-Growth algorithm includes two main steps:\begin{enumerate}
    \item FP-Tree Construction: Initially, the database is scanned to count item frequencies, discarding those under a set minimum support. The dataset is scanned again to construct the FP-Tree, inserting transactions so that shared items appear along branches. Nodes symbolize items, and paths represent itemsets, with node linkage enabling efficient tree traversal.
    \item Extracting Frequent Itemsets: After forming the FP-Tree, frequent itemsets are mined by recursively identifying patterns, starting with the least frequent items to form conditional FP-Trees. These conditional FP-Trees focus on paths with a specific item, recursively breaking down and solving the problem.
\end{enumerate}Detailed explanation of the algorithmic steps involved in FP-Growth are available in \cite{han2000mining}.

\section{Application of the framework - Maximum Clique Problem}
\label{sec:application_of_framework_mcp}

The GCO-HPIF framework is now applied to the maximum clique problem (MCP), a graph-based COP to demonstrate its performance. 

\subsection{Maximum Clique Problem}
In MCP, the objective is to identify the largest complete subgraph within a given graph \cite{wu2015review}. 
The decision variant of the MCP is renowned for being one of the foundational NP-complete problems identified by \cite{Karp1972}. MCP has substantial applications across various fields, including drug design \cite{konc2022protein}, telecommunications \cite{phillipson2023quantum}, bioinformatics \cite{wang2022mini}, social network analysis \cite{cavique2018data}, and integrated circuit design \cite{azriel2019sok}. MCP has been extensively studied owing to its theoretical significance and numerous practical uses. Formally, it is defined as follows: Consider an undirected, unweighted graph $G = (V,E)$ where $V$ represents the node set, and $E$ represents the edge set. The vertex cardinality, denoted $|V|$, is the count of vertices, while edge cardinality, denoted $|E|$, is the edge count. For a node subset $S \subseteq V$, the induced subgraph is $G(S) = G(S,E \subseteq S \times S)$. A graph is complete if each vertex pair is connected, i.e., $\forall (i,j) \in V$, if $i \neq j$, then $(i,j) \in E$. A clique $C$ is a complete subgraph of $G$. A clique is maximal if it is not contained in any larger clique within $G$. A maximum clique has the largest vertex cardinality of all cliques in the graph. The MCP can be expressed as the following integer linear programming problem:

\begin{equation}
\label{eq:mcp_ilp_formulation}
\begin{aligned}
    \text{Maximize } & \sum_{i \in V} x_i \\
    \text{subject to: } & x_i + x_j \leq 1 \quad \forall (i, j) \notin E \\
    & x_i \in \{0, 1\} \quad \forall i \in V
\end{aligned}
\end{equation}

In this context, \( x_i \) is a binary variable indicating the inclusion (\(x_i = 1\)) or exclusion (\(x_i = 0\)) of vertex \( i \) in the clique. The inequality constraint detailed in Equation~\ref{eq:mcp_ilp_formulation} requires that vertices linked by non-existent edges cannot both be part of a clique. The clique number \( \omega(G) \) is defined as the total number of vertices in the largest clique within \( G \). Besides this traditional maximum clique problem (MCP) on unweighted graphs, there are weighted variants. If each vertex \( i \) carries a weight \( w_i \), these can be organized into a weight vector \( \vec{w} \in \mathbb{R}^{N} \). The maximum weighted clique problem (MWCP) seeks a clique optimizing the sum of node weights, and can likewise consider edge weights. This paper focuses on the classic MCP. Numerous reviews \cite{pardalos1994maximum,bomze1999maximum,wu2015review} and studies \cite{feige2004approximating,engebretsen2003towards,zuckerman2006linear,chalermsook2020gap} have explored the hardness of MCP. However, no prior work has investigated MCP instance hardness via graph features using machine learning, and the literature lacks studies on explaining machine learning-based predictions of MCP instance hardness. Thus, MCP serves as the graph-based COP in evaluating the GCO-HPIF framework.

\subsection{Algorithms}

Multiple algorithmic strategies to tackle the MCP have been developed, encompassing exact methods like branch and bound (B\&B) \cite{li2017minimization}, \cite{li2010efficient}, \cite{jiang2016combining}, \cite{san2019new}, \cite{san2023clisat}, greedy and local search techniques \cite{wu2012multi}, \cite{wang2016two}, \cite{wang2020sccwalk}, Monte Carlo methods \cite{angelini2021mismatching}, and more recently, approaches using graph neural networks (GNN) \cite{karalias2020erdos}, \cite{min2022can}, \cite{sanokowski2024variational} and quantum computing \cite{haverly2021implementation}, \cite{li2024parameter}, \cite{pelofske2019solving}. This study chooses five algorithms: Gurobi's Mixed Integer Linear Programming Solver (Gurobi) \cite{gurobi}, CliSAT \cite{san2023clisat}, Mixed Order Maximum Clique (MOMC) \cite{li2017minimization}, Erdos goes Neural (EGN) \cite{karalias2020erdos}, and Hybrid Geometric Scattering (HGS) \cite{min2022can}. Gurobi, CliSAT, and MOMC are established exact B\&B-based algorithms, while EGN and HGS represent SOTA unsupervised GNN-based techniques for the MCP. Table \ref{tab:summary_of_mcp_algorithms} provides an overview of these algorithms utilized for the MCP.

\begin{table}[h]
    \centering
    \caption{Summary of algorithms used to demonstrate the GCO-HPIF framework for the case of maximum clique problem}
    \begin{tabularx}{\textwidth}{|l|X|X|}
        \hline
        \textbf{Algorithm/Solver} & \textbf{Exact} & \textbf{Mechanism of Computation} \\
        \hline
        Gurobi & Yes & Branch and Cut \\
        \hline
        CliSAT & Yes & Branch and Bound \\
        \hline
        Mixed ordering MaxClique Solver (MOMC) & Yes & Branch and Bound \\
        \hline
        Erdos Goes Neural (EGN) & No & Graph Neural Networks \\
        \hline
        Hybrid Geometric Scattering (HGS) & No & Graph Neural Networks \\
        \hline
    \end{tabularx}
    \label{tab:summary_of_mcp_algorithms}
\end{table}

\subsection{Dataset}
A total of 6135 graph instances were gathered for analysis, with a summary presented in Table \ref{tab:summary_of_graph_dataset}. We chose TWITTER, COLLAB, and IMDB-BINARY datasets due to their frequent application in assessing GNN-based MCP algorithms in the literature.

\begin{table}[h]
    \centering
    \caption{Summary of graph instances datasets used for hardness prediction and explanation for the maximum clique problem}
    \begin{tabularx}{\textwidth}{>{\centering\arraybackslash}X >{\centering\arraybackslash}X >{\centering\arraybackslash}X}
        \hline
        \textbf{Dataset Name} & \textbf{Source} & \textbf{Number of Graphs} \\
        \hline
        TWITTER & \cite{yan2008mining} & 140 \\
        \hline
        COLLAB & \cite{yanardag2015deep} & 4996 \\
        \hline
       IMDB-BINARY & \cite{yanardag2015deep} & 999 \\
        \hline
    \end{tabularx}
    \label{tab:summary_of_graph_dataset}
\end{table}

Of the 6135 graph instances, Gurobi and CliSAT resolved them all accurately. All five algorithms successfully solved 2906 instances, whereas 3287 graphs remained unsolved by at least one of the three algorithms: MOMC, EGN, or HGS. Figure \ref{fig:graphs_for_predicting_instance_hardness_mcp} displays the node, edge, and density distributions of these 3287 graphs.

\afterpage{%
\begin{figure}[H]
    \centering
    \includegraphics[width=0.9\textwidth]{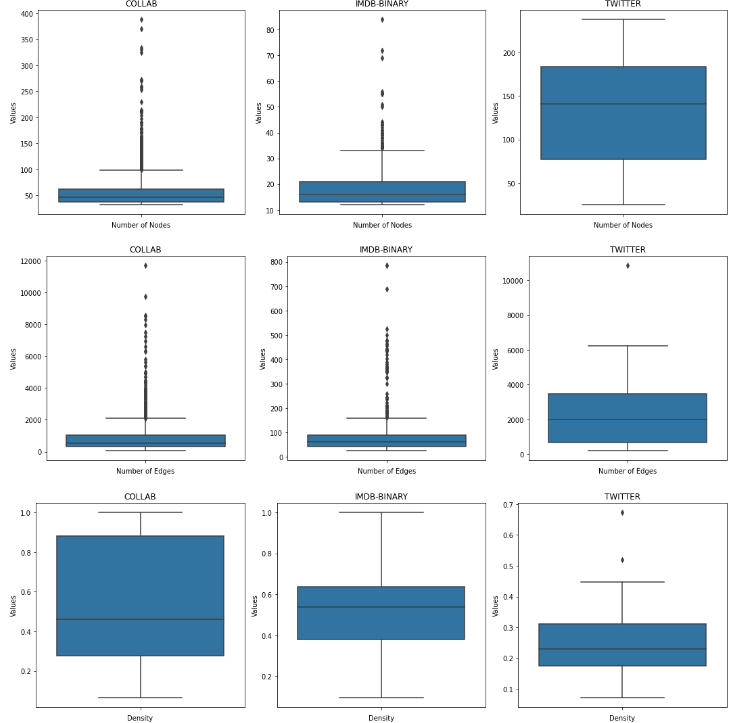}
    \caption{Node count, edge count, and density distribution of 3287 graph instances used for predicting instance hardness of the MCP}
    \label{fig:graphs_for_predicting_instance_hardness_mcp}
\end{figure}
}

\subsection{Hardness measure}

A straightforward measure of difficulty was applied to the dataset consisting of the graphs depicted in Figure \ref{fig:graphs_for_predicting_instance_hardness_mcp}. If an algorithm successfully identified the maximum clique of a graph, that graph was not deemed difficult for the algorithm. With Gurobi and CliSAT determining the maximum cliques for all 6135 graph instances, including the 3287 mentioned in Figure \ref{fig:graphs_for_predicting_instance_hardness_mcp}, this difficulty measure pertains solely to EGN, HGS, and MOMC. Consequently, a feasible hardness criterion was established: an instance was labeled as hard if none of these three algorithms could find a maximum clique; otherwise, it was not. Following this criterion, 111 instances were identified as hard, while 3176 were not. Specifically, the maximum clique sizes of 111 instances eluded all three algorithms, whereas for the other 3176 instances, at least one or two algorithms succeeded in identifying the sizes. The measure that we chose may seem to be simplistic as it constitutes a binary and absolute measure without considering the size of the clique found (for example, if an algorithm finds a clique size of 90\% the size of the maximum clique for an instance, the instance is still classified as hard for that algorithm) and does not incorporate computational time into the hardness measure. However, this simplicity allows for identifying instances for which several algorithms were simultaneously unable to find the maximum cliques. Also, it works well for those situations where finding the optimal (and not just near optimal) solutions is important.    

\subsection{Computational considerations}

The computations were performed on a Linux Ubuntu 22.04.3 LTS, 64-bit with an AMD Ryzen 1920 processor (24 threads, 4GHz CPU speed) and an 8GB TU106 GeForce RTX 2070 GPU. Version 10 of the Gurobi solver was used and NetworkX's version 2.8.4 was used for computing graph features. CliSAT, MOMC, and HGS algorithms necessitated graph format conversion, with pre-processing time excluded from run-time calculations. For CliSAT, the maximum clique computation time was determined by summing the times $ts$, $tr$, and $tp$ from output files, though $ts$ dominated in most cases. MOMC showed slight variability in run times for identical instances, prompting six runs per instance. The median of these runs was recorded as the runtime for MOMC.

\subsection{Machine learning based hardness prediction methodology}

Figure \ref{fig:ml_model_development_instance_hardness_prediction} illustrates the methodology for an ML-based instance hardness prediction model. Given the dataset's significant imbalance, with only 3.3\% hard instances, we applied stratified sampling to divide the data into 80\% training and 20\% testing sets. A min-max scaler normalized feature values to mitigate their disproportionate effect on model predictions. We employed three classifiers: support vector classifier (SVC), logistic regression (LR), and XGBoost (XGB), tuning their hyperparameters. Each model underwent five-fold cross-validation, using the weighted F1-score to evaluate performance for both classes. The classifier with the highest weighted F1-score was selected for predicting instance hardness.

\section{Results and Discussion}
\label{sec:results_and_discussion}

\subsection{Instance hardness prediction results}

Table \ref{tab:best_svc_xgboost_lr_models_after_training} presents the outcomes of the best XGB, SVC, and LR models. Among them, XGB achieved the highest ROC-AUC score, while SVC attained the highest weighted and minority class F1-scores. 

\begin{table}[h]
    \centering
    \caption{Summary of the best performing XGB, SVC and LR models for instance hardness prediction for the case of MCP. In the confusion matrix column, TN, FP, FN and TP are abbreviations of true negative, false positive, false negative and true positive respectively. Negative (0) and positive (1) in this context stand for not hard and hard problem instances respectively.}
    \begin{tabularx}{\textwidth}{|p{3cm}|p{2cm}|p{3cm}|p{2cm}|X|} 
        \hline
        \textbf{Model} & \textbf{Weighted \newline F1-Score} & \textbf{Class-wise F1-Score} & \textbf{ROC-AUC Score} & \textbf{Confusion Matrix} \\
        \hline
        XGBoost & 0.9908 & [0.9952,0.8636] & 0.9294 &  [[TN=633,FP=3],\newline [FN=3,TP=19]] \\
        \hline
        Logistic \newline Regression & 0.9906 & [0.9952,0.8571] & 0.9075 & [[TN=634,FP=2],\newline [FN=4,TP=18]] \\
        \hline
        Support \newline Vector \newline Classifier & 0.9921 & [0.9960,0.8780] & 0.9083 & [[TN=635,FP=1],\newline [FN=4,TP=18]] \\
        \hline
    \end{tabularx}
    \label{tab:best_svc_xgboost_lr_models_after_training}
\end{table}

\begin{figure}[h]
    \centering
    \includegraphics[width=\textwidth]{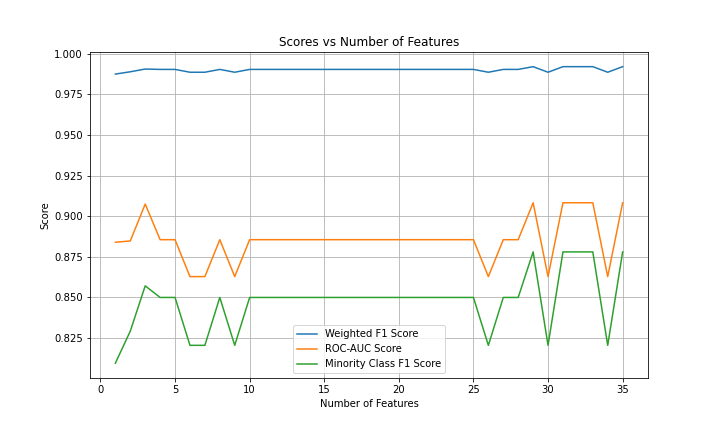}
    \caption{Model performance metrics vs number of features for the SVC model}
    \label{fig:performance_measures_vs_number_of_features_svc}
\end{figure}

Figure \ref{fig:performance_measures_vs_number_of_features_svc} illustrates the variation of three key model performance metrics—weighted F1-score, ROC-AUC score, and minority class F1-score—based on the number of features for the SVC model. The model attains optimal weighted F1-score and ROC-AUC score with just 3 features. Although the minority class F1-score slightly improves with more features, 3 features are favored for better explainability, given the marginal performance gain with 30 features. Thus, the SVC model with 3 features, specifically the second smallest and smallest eigenvalues of the adjacency matrix, and the number of nodes, was chosen as the instance hardness prediction model.

\subsection{Hardness prediction explanation}

Table \ref{tab:association-mining-rules-fp-growth} presents the association rules identified by the FP-Growth algorithm for the instance hardness dataset. To prevent overfitting and excessively complex rules, the antecedent support threshold was set to 10\%, ensuring rules reflect trends in at least 10\% of the dataset. Given that hard instances account for just 3\% of the dataset, the FP-Growth algorithm was applied specifically to these cases, to accelerate the generation of association rules. However, the validity of the generated rules was subsequently verified by testing instance compliance with these rules and their associated hardness for the non-hard instances too. An important rule is detailed in Expression \ref{eq:key_association_mining_rule_mcp_hardness} (where P25 indicates the 25th percentile, etc.): 

\begin{equation}
    \begin{aligned}[t]
        &\text{P0 value} < \text{Second Smallest Eigenvalue of Adjacency Matrix} < \text{P25 value}, \\
        &\text{P0 value} < \text{Smallest Eigenvalue of Adjacency Matrix} < \text{P25 value}, \\
        &\text{P75 value} < \text{Number of Nodes} < \text{P100 value}
    \end{aligned}
    \label{eq:key_association_mining_rule_mcp_hardness}
\end{equation}\\

Figure \ref{fig:AR_Nodes_75_100_SEA_0_25_SSEA_0_25} reveals that this association rule is applicable to 88.29\% of hard instances and inapplicable to 87.62\% of non-hard instances. This rule effectively explains machine learning algorithm predictions with high accuracy and distinguishes between hardness classes. After feature selection, the number of nodes, smallest eigenvalue adjacency, and second smallest eigenvalue adjacency are notably interrelated, as shown in Figure \ref{fig:correlation_plot_three_features_instance_hardness}. These correlated features frequently appear in the association rules listed in Table \ref{tab:association-mining-rules-fp-growth}. Among the seven rules, the rule in Expression \ref{eq:key_association_mining_rule_mcp_hardness} proficiently discriminates between hard and non-hard classes, making it the most explainable rule for this graph-based COP and dataset. Although some rules like Expression \ref{eq:association_mining_rule1_mcp_hardness} achieve higher accuracy on hard instances (e.g., 94.6\%), they perform worse on non-hard instances (e.g., 77.4\%) compared to the rule in Expression \ref{eq:key_association_mining_rule_mcp_hardness}.

\begin{equation}
    \begin{aligned}[t]
        &\text{P0 value} < \text{Second Smallest Eigenvalue of Adjacency Matrix} < \text{P25 value}
    \end{aligned}
    \label{eq:association_mining_rule1_mcp_hardness}
\end{equation}\\

The seven association rules identified by FP-Growth in Table \ref{tab:association-mining-rules-fp-growth} possess high support, with the lowest support being 0.8829, confirming their validity for most hard instances. Notably, the lift value for these rules was consistently 1, due to the focus on hard instances alone, leading to equal numerator and denominator values in lift calculation. In datasets with a higher proportion of hard instances exceeding a 10\% threshold, FP-Growth can uncover rules with lift values surpassing 1. While several rules from this dataset achieved lift values near 4, they were not beneficial for understanding hardness, as the consequents identified included other graph features or combinations rather than standalone hardness classes.    

\afterpage{%
\begin{table}[h]
    \centering
    \caption{Association rules derived from the application of FP-Growth algorithm to the instance hardness dataset. Consequent support, confidence and lift values are 1 for all the rules. Therefore, antecedent support is the same as support for all the rules.}
    \begin{tabularx}{\textwidth}{|>{\centering\arraybackslash}p{8.5cm}|>{\centering\arraybackslash}p{2.25cm}|>
    {\centering\arraybackslash}p{1.25cm}|>
    {\centering\arraybackslash}X|} 
        \hline
        \textbf{Antecedents} & \textbf{Consequents \newline (Problem Hardness)} & \textbf{Support} & \textbf{Overall \newline Accuracy} \\
        \hline
         $75th \, percentile \, value <$ Number of Nodes $< 100th \, percentile \, value$ & Hard & 0.9009 & 77.51\% \\
         \hline
         $0th \, percentile \, value <$ Smallest Eigenvalue Adjacency $< 25th \, percentile \, value$  & Hard & 
        0.9099 & 77.76\% \\
        \hline
        $0th \, percentile \, value <$ Second Smallest Eigenvalue Adjacency $<25th \, percentile \, value$   & Hard &
        0.9459 & 78\% \\
        \hline
        $0th \, percentile \, value <$ Smallest Eigenvalue Adjacency $< 25th \, percentile \, value$, \newline $0th \, percentile \, value <$ Second Smallest Eigenvalue Adjacency $<25th \, percentile \, value$  & Hard & 0.9009 & 83.69\% \\
        \hline
        $75th \, percentile \, value <$ Number of Nodes $< 100th \, percentile \, value$, 
        \newline $0th \, percentile \, value <$ Second Smallest Eigenvalue Adjacency $<25th \, percentile \, value$  & Hard & 0.9009 & 85.70\% \\
        \hline
        $75th \, percentile \, value <$ Number of Nodes $< 100th \, percentile \, value$, 
        \newline $0th \, percentile \, value <$ Smallest Eigenvalue Adjacency $< 25th \, percentile \, value$  & Hard & 0.8829 & 86.06\% \\
        \hline
         $75th \, percentile \, value <$ Number of Nodes $< 100th \, percentile \, value$, 
        \newline $0th \, percentile \, value <$ Smallest Eigenvalue Adjacency $< 25th \, percentile \, value$
        \newline $0th \, percentile \, value <$ Second Smallest Eigenvalue Adjacency $<25th \, percentile \, value$
        & Hard & 0.8829 & 87.64\% \\
        \hline     
    \end{tabularx}
    \label{tab:association-mining-rules-fp-growth}
\end{table}
}

\afterpage{%
\begin{figure}[h]
    \centering
    \includegraphics[width=0.65\textwidth]{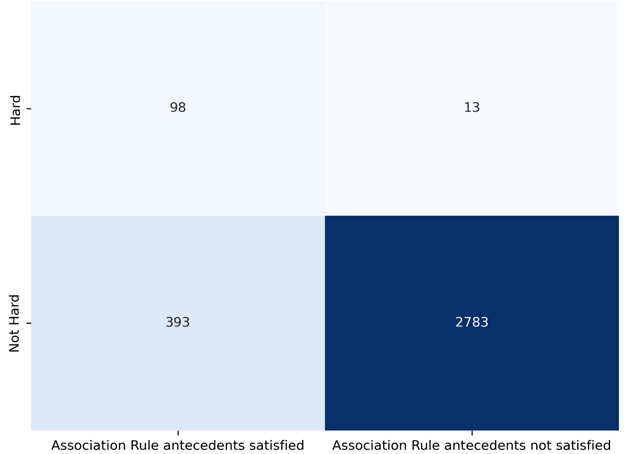}
    \caption{Relationship between association rule specified in Expression \ref{eq:key_association_mining_rule_mcp_hardness} and instance hardness}
    \label{fig:AR_Nodes_75_100_SEA_0_25_SSEA_0_25}
\end{figure}
}

\afterpage{%
\begin{figure}[h]
    \centering
    \includegraphics[width=\textwidth]{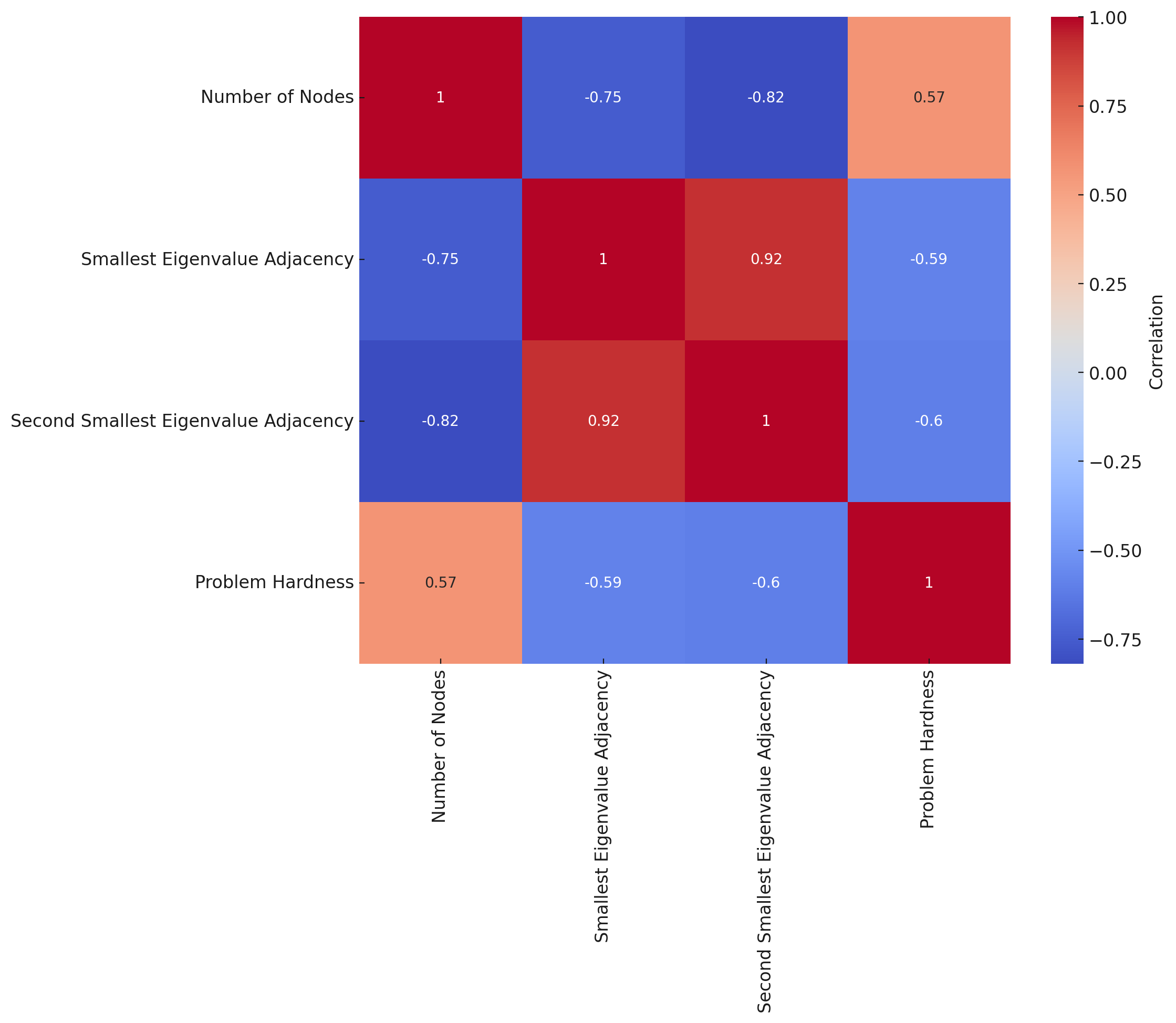}
    \caption{Correlation heatmap depicting the correlation of the three features - Second Smallest Eigenvalue of Adjacency Matrix, Smallest Eigenvalue of Adjacency Matrix and Number of Nodes with each other and with the output variable, Instance Hardness}
    \label{fig:correlation_plot_three_features_instance_hardness}
\end{figure}
}

\subsection{Computation time prediction}
A machine learning-based regression model was created to forecast the computation time of MCP instances, alongside predicting their hardness. Regression models using XGB and RF were trained on a dataset of 3287 graphs for each of the five algorithms. Much like the approach for instance hardness prediction, this dataset was split into training and test sets with an 80:20 ratio, followed by feature scaling via a min-max scaler to mitigate the impact of large feature values. Both XGB and RF regressors underwent five-fold cross-validation, targeting the negative root mean squared error (RMSE) as the objective function during hyperparameter tuning. It emerged that the RF model provided slightly better computation time predictions than XGB, although both models performed similarly in metrics. Notably, predictive models performed well in terms of high $R^{2}$ and low RMSE percentages for GNN-based solvers HGS and EGN, yet underperformed with exact B\&B algorithms such as Gurobi, MOMC, and CliSAT. This disparity could be due to factors such as the three to five orders of magnitude spanned by computation times for MOMC, CliSAT, and Gurobi, as opposed to just one order for EGN and HGS. Exact algorithms tend to scale exponentially with problem size, unlike the linear scaling of GNN-based methods, complicating accurate prediction for B\&B algorithms. Additionally, in CliSAT's case, computation time increased by 0.05 seconds beyond a certain problem size, shown by the dense patch at 0.05 seconds in Figure \ref{fig:computation_time_algorithms}. Also, noticeable outliers in Gurobi and MOMC may significantly skew prediction accuracy. 

\afterpage{%
\begin{figure}[H]
    \centering
    \includegraphics[width=0.75\textwidth]{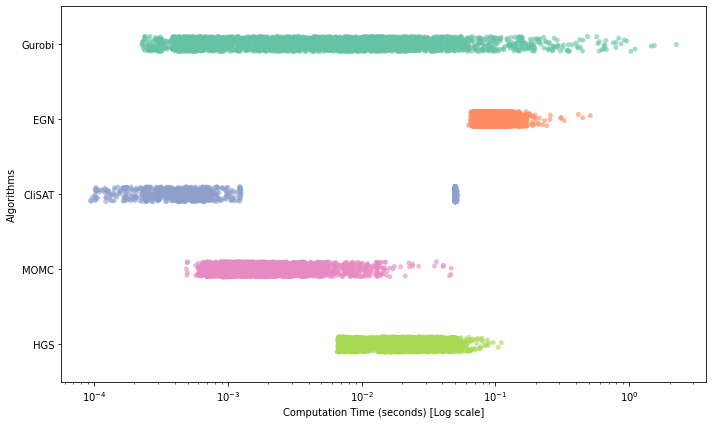}
    \caption{Computation time distribution of the five algorithms namely Gurobi, EGN, CliSAT, MOMC and HGS for the 3287 MCP instances plotted on a logarithmic scale}
    \label{fig:computation_time_algorithms}
\end{figure}

Table \ref{tab:computation_time_prediction_results} presents the outcomes of XGB and RF regression models applied to 658 test instances of the MCP using GNN-based algorithms EGN and HGS. The percentage RMSE and $R^2$ values for both XGB and RF are similar across HGS and EGN. Prediction of computation time for HGS surpasses that for EGN in each model. An $R^2$ value above 0.75 across all scenarios suggests that the features of the dataset account for most of the variance in computation time.           
}

\begin{table}[h]
\centering
\caption{Summary of the performance XGB and RF based regression models for predicting the computation time of GNN-based EGN and HGS algorithms for 658 MCP test instances}

\begin{tabular}
{|p{3cm}|p{2cm}|p{3cm}|p{2cm}|} 
    \hline
    \textbf{Regression \newline Model} & \textbf{Algorithm} & \textbf{Percentage RMSE} & \textbf{\( R^2 \)}\\
    \hline 
    RF & HGS & 4.55 & 0.993 \\
    \hline
    XGB & HGS & 5.12 & 0.991 \\
    \hline
    RF & EGN & 13.07 & 0.771 \\
    \hline
    XGB & EGN & 14.33 & 0.725 \\
    \hline

\end{tabular}
\label{tab:computation_time_prediction_results}

\end{table}

\begin{figure}[H]
    \centering
    \includegraphics[width=\textwidth]{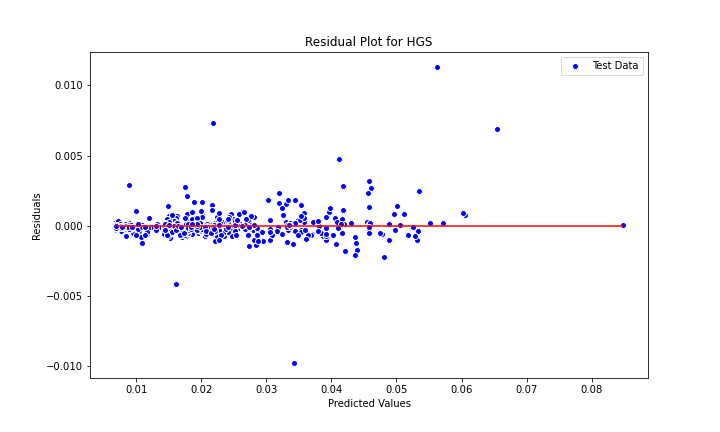}
    \caption{Residuals plot of the 658 test instances for RF regression model for HGS computation time prediction}
    \label{fig:residual_plot_HGS_computation_time_prediction}
\end{figure}

Figure \ref{fig:residual_plot_HGS_computation_time_prediction} displays the residuals for 658 test cases of the RF regression model predicting HGS computation time. The plot reveals that the residuals predominantly center around 0, suggesting a robust fit.   

According to prediction algorithms, employing problem-agnostic graph-based features effectively predicts computation time for GNN-based algorithms. However, further improvement is needed for this method to be applicable to predicting B\&B-based exact algorithms' computation times.   

\section{Conclusion and further work}
\label{sec:conclusion_and_further_research}

GCO-HPIF is a comprehensive framework designed to assess and explain the difficulty of graph-based COPs. It relies on generic graph features to predict instance hardness by employing ML-based classification algorithms trained on datasets comprising these features and hardness labels. For explanation, the FP-Growth association rule mining algorithm is utilized to identify feature combinations that render an instance challenging. The application of GCO-HPIF to the MCP, a well-known NP-hard problem, involved computing 23 graph features for a dataset of 3287 graphs sourced from COLLAB, IMDB, and TWITTER datasets. Maximum cliques were evaluated using five algorithms, namely, the exact Branch and Bound (B\&B) algorithms (Gurobi, CliSAT, and EGN) and GNN-based algorithms (EGN and HGS). Gurobi and CliSAT successfully computed maximum cliques for all instances, whereas MOMC, EGN, and HGS were unable to do so for 111 instances, which were then labelled as hard. These instances were easily identified using a Support Vector Classifier, which achieved a weighted F1-score of 0.9921, a minority F1-score of 0.878, and an ROC-AUC of 0.9083, based on only three graph features: Second Smallest and Smallest Eigenvalues of the Adjacency Matrix, and the Number of Nodes. The FP-Growth algorithm associated hardness predictions with feature quartiles, achieving an 87.64\% accuracy. For predicting algorithm runtime on MCP instances, RF and XGB regression models were employed, with the best RF model achieving a 5.12\% RMSE and an $R^2$ of 0.991 for HGS. \\

The results of predicting hardness, explanation, and computational time using generic graph-based features for an NP-hard problem suggest that the proposed framework holds potential to become a general approach for predicting and understanding hardness, as well as predicting computation times in graph-based COP algorithms. However, challenges remain that open avenues for further investigation. Firstly, the framework needs assessment across a broader range of graph-based COPs. Tasks involving edge information and path-based objectives like TSP and LRP necessitate additional graph-based features that capture edge data, complementing those used in this study. Evaluating the framework on TSP and LRP could provide deeper insights into its capabilities and limitations. Secondly, the largest graph examined contained 388 nodes, while graphs with thousands of nodes exist, necessitating an examination of the framework’s scalability. This can also provide further insights into whether the same association rules still hold good for these larger graphs or are a different set of association rules found for them. Thirdly, applying different criteria for measuring hardness could reveal variations in hardness predictions and explanations. For instance, a hardness metric combining clique size with computation time could identify features making instances challenging for algorithms such as Gurobi and CliSAT, which faced no hard instances in this dataset. Lastly, improving the prediction performance for exact B\&B algorithms' computation time model can be another research direction. This could involve transforming the computation time variable to a different scale such as a logarithmic scale and retraining, or engaging in refined feature engineering and feature selection of graph-based features.\\

\textbf{CRediT authorship contribution statement} \\
\textbf{Bharat S. Sharman}: Conceptualization, Methodology, Software, Validation,
Formal analysis, Investigation, Data curation, Writing – original
draft, Writing – review \& editing, Visualization. \textbf{Elkafi Hassini}: Conceptualization,
Methodology, Project administration, Resources, Supervision, Writing – review
\& editing.\\

\textbf{Compliance with Ethical Standards}: 
\begin{enumerate}
    \item \textbf{Funding}: The authors did not receive support from any organization for the submitted work.
    \item \textbf{Conflict of Interest}:  Bharat Sharman declares that he has no conflict of interest. Elkafi Hassini declares that has no conflict of interest.
    \item \textbf{Interests}: The authors have no relevant financial or non-financial interests to disclose.
    \item \textbf{Ethical Approval}: This article does not contain any studies with human participants or animals performed by any of the authors.
\end{enumerate}

\textbf{Data Availability}
 The datasets used for this study are publicly available and can be freely downloaded from the websites of the cited sources.  

\textbf{Acknowledgments}

The authors express gratitude to Gurobi Optimization, LLC for granting free academic license access. Thanks are also extended to the creators of the CliSAT, MOMC, EGN \& HGS algorithms, whose source code, shared with or made accessible to the authors, was crucial for this study.\\

\textbf{Declaration of generative AI and AI-assisted technologies in the writing process}

During the preparation of this work, the author(s) used Writefull for scientific writing to paraphrase their initial draft to improve its scientific language. After using this tool, the authors reviewed and edited the content as needed and take full responsibility for the content of the published article.




\bibliography{cas_refs}

@book{du2022introduction,
  title={Introduction to Combinatorial Optimization},
  author={Du, Dingzhu and Pardalos, Panos M and Hu, Xiaodong and Wu, Weili and others},
  year={2022},
  publisher={Springer}
}

@article{pop2024comprehensive,
  title={A comprehensive survey on the generalized traveling salesman problem},
  author={Pop, Petric{\u{a}} C and Cosma, Ovidiu and Sabo, Cosmin and Sitar, Corina Pop},
  journal={European Journal of Operational Research},
  volume={314},
  number={3},
  pages={819--835},
  year={2024},
  publisher={Elsevier}
}

@article{zhuang2024improving,
  title={Improving order picking efficiency through storage assignment optimization in robotic mobile fulfillment systems},
  author={Zhuang, Yanling and Zhou, Yun and Hassini, Elkafi and Yuan, Yufei and Hu, Xiangpei},
  journal={European Journal of Operational Research},
  volume={316},
  number={2},
  pages={718--732},
  year={2024},
  publisher={Elsevier}
}

@article{mansouri2019optimal,
  title={Optimal pricing in iterative flexible combinatorial procurement auctions},
  author={Mansouri, Bahareh and Hassini, Elkafi},
  journal={European Journal of Operational Research},
  volume={277},
  number={3},
  pages={1083--1097},
  year={2019},
  publisher={Elsevier}
}

@article{calvet2023role,
  title={On the role of metaheuristic optimization in bioinformatics},
  author={Calvet, Laura and Benito, Sergio and Juan, Angel A and Prados, Ferran},
  journal={International Transactions in Operational Research},
  volume={30},
  number={6},
  pages={2909--2944},
  year={2023},
  publisher={Wiley Online Library}
}

@article{ren2023review,
  title={A review of combinatorial optimization problems in reverse logistics and remanufacturing for end-of-life products},
  author={Ren, Yaping and Lu, Xinyu and Guo, Hongfei and Xie, Zhaokang and Zhang, Haoyang and Zhang, Chaoyong},
  journal={Mathematics},
  volume={11},
  number={2},
  pages={298},
  year={2023},
  publisher={MDPI}
}

@incollection{li2023scheduling,
  title={Scheduling of Semiconductor Manufacturing System},
  author={Li, Li and Yu, Qingyun and Lin, Kuo-Yi and Ma, Yumin and Qiao, Fei},
  booktitle={Data-Driven Scheduling of Semiconductor Manufacturing Systems},
  pages={1--24},
  year={2023},
  publisher={Springer}
}

@inproceedings{zhang2022learning,
  title={Learning to solve travelling salesman problem with hardness-adaptive curriculum},
  author={Zhang, Zeyang and Zhang, Ziwei and Wang, Xin and Zhu, Wenwu},
  booktitle={Proceedings of the AAAI Conference on Artificial Intelligence},
  volume={36},
  pages={9136--9144},
  year={2022}
}

@article{mersmann2013novel,
  title={A novel feature-based approach to characterize algorithm performance for the traveling salesperson problem},
  author={Mersmann, Olaf and Bischl, Bernd and Trautmann, Heike and Wagner, Markus and Bossek, Jakob and Neumann, Frank},
  journal={Annals of Mathematics and Artificial Intelligence},
  volume={69},
  pages={151--182},
  year={2013},
  publisher={Springer}
}

@article{cardenas2018creating,
  title={Creating hard-to-solve instances of travelling salesman problem},
  author={C{\'a}rdenas-Montes, Miguel},
  journal={Applied Soft Computing},
  volume={71},
  pages={268--276},
  year={2018},
  publisher={Elsevier}
}

@article{cvetkovic2018complexity,
  title={Complexity indices for the traveling salesman problem based on short edge subgraphs},
  author={Cvetkovi{\'c}, Drago{\v{s}} and {\v{C}}angalovi{\'c}, Mirjana and Dra{\v{z}}i{\'c}, Zorica and Kova{\v{c}}evi{\'c}-Vuj{\v{c}}i{\'c}, Vera},
  journal={Central European Journal of Operations Research},
  volume={26},
  pages={759--769},
  year={2018},
  publisher={Springer}
}

@article{hoos2014empirical,
  title={On the empirical scaling of run-time for finding optimal solutions to the travelling salesman problem},
  author={Hoos, Holger H and St{\"u}tzle, Thomas},
  journal={European Journal of Operational Research},
  volume={238},
  number={1},
  pages={87--94},
  year={2014},
  publisher={Elsevier}
}

@article{heins2023study,
  title={A study on the effects of normalized TSP features for automated algorithm selection},
  author={Heins, Jonathan and Bossek, Jakob and Pohl, Janina and Seiler, Moritz and Trautmann, Heike and Kerschke, Pascal},
  journal={Theoretical Computer Science},
  volume={940},
  pages={123--145},
  year={2023},
  publisher={Elsevier}
}

@article{huerta2022improving,
  title={Improving the state-of-the-art in the traveling salesman problem: An anytime automatic algorithm selection},
  author={Huerta, Isa{\'\i}as I and Neira, Daniel A and Ortega, Daniel A and Varas, Vicente and Godoy, Julio and Asin-Acha, Roberto},
  journal={Expert Systems with Applications},
  volume={187},
  pages={115948},
  year={2022},
  publisher={Elsevier}
}

@inproceedings{seiler2020deep,
  title={Deep learning as a competitive feature-free approach for automated algorithm selection on the traveling salesperson problem},
  author={Seiler, Moritz and Pohl, Janina and Bossek, Jakob and Kerschke, Pascal and Trautmann, Heike},
  booktitle={International Conference on Parallel Problem Solving from Nature},
  pages={48--64},
  year={2020},
  organization={Springer}
}

@inproceedings{zhao2021towards,
  title={Towards feature-free TSP solver selection: A deep learning approach},
  author={Zhao, Kangfei and Liu, Shengcai and Yu, Jeffrey Xu and Rong, Yu},
  booktitle={2021 International Joint Conference on Neural Networks (IJCNN)},
  pages={1--8},
  year={2021},
  organization={IEEE}
}

@inproceedings{pihera2014application,
  title={Application of machine learning to algorithm selection for TSP},
  author={Pihera, Josef and Musliu, Nysret},
  booktitle={2014 IEEE 26th International Conference on Tools with Artificial Intelligence},
  pages={47--54},
  year={2014},
  organization={IEEE}
}

@article{cricsan2021randomness,
  title={On Randomness and Structure in Euclidean TSP Instances: A Study With Heuristic Methods},
  author={Cri{\c{s}}an, Gloria Cerasela and Nechita, Elena and Simian, Dana},
  journal={IEEE Access},
  volume={9},
  pages={5312--5331},
  year={2021},
  publisher={IEEE}
}

@inproceedings{lindauer2018warmstarting,
  title={Warmstarting of model-based algorithm configuration},
  author={Lindauer, Marius and Hutter, Frank},
  booktitle={Proceedings of the AAAI Conference on Artificial Intelligence},
  volume={32},
  pages={1355--1362},
  year={2018}
}

@inproceedings{cameron2020predicting,
  title={Predicting propositional satisfiability via end-to-end learning},
  author={Cameron, Chris and Chen, Rex and Hartford, Jason and Leyton-Brown, Kevin},
  booktitle={Proceedings of the AAAI Conference on Artificial Intelligence},
  volume={34},
  pages={3324--3331},
  year={2020}
}

@article{chang2022predicting,
  title={Predicting propositional satisfiability based on graph attention networks},
  author={Chang, Wenjing and Zhang, Hengkai and Luo, Junwei},
  journal={International Journal of Computational Intelligence Systems},
  volume={15},
  number={1},
  pages={84},
  year={2022},
  publisher={Springer}
}

@article{jooken2022new,
  title={A new class of hard problem instances for the 0--1 knapsack problem},
  author={Jooken, Jorik and Leyman, Pieter and De Causmaecker, Patrick},
  journal={European Journal of Operational Research},
  volume={301},
  number={3},
  pages={841--854},
  year={2022},
  publisher={Elsevier}
}

@article{smith2021revisiting,
  title={Revisiting where are the hard knapsack problems? via instance space analysis},
  author={Smith-Miles, Kate and Christiansen, Jeffrey and Mu{\~n}oz, Mario Andr{\'e}s},
  journal={Computers \& Operations Research},
  volume={128},
  pages={105184},
  year={2021},
  publisher={Elsevier}
}

@article{smith2014exploring,
  title={Exploring the role of graph spectra in graph coloring algorithm performance},
  author={Smith-Miles, Kate and Baatar, Davaatseren},
  journal={Discrete Applied Mathematics},
  volume={176},
  pages={107--121},
  year={2014},
  publisher={Elsevier}
}

@article{smith2012measuring,
  title={Measuring instance difficulty for combinatorial optimization problems},
  author={Smith-Miles, Kate and Lopes, Leo},
  journal={Computers \& Operations Research},
  volume={39},
  number={5},
  pages={875--889},
  year={2012},
  publisher={Elsevier}
}

@article{san2023clisat,
  title={CliSAT: A new exact algorithm for hard maximum clique problems},
  author={San Segundo, Pablo and Furini, Fabio and {\'A}lvarez, David and Pardalos, Panos M},
  journal={European Journal of Operational Research},
  volume={307},
  number={3},
  pages={1008--1025},
  year={2023},
  publisher={Elsevier}
}

@article{wolpert1997no,
  title={No free lunch theorems for optimization},
  author={Wolpert, David H and Macready, William G},
  journal={IEEE transactions on evolutionary computation},
  volume={1},
  number={1},
  pages={67--82},
  year={1997},
  publisher={IEEE}
}

@article{khanna2023distinctive,
  title={A distinctive explainable machine learning framework for detection of polycystic ovary syndrome},
  author={Khanna, Varada Vivek and Chadaga, Krishnaraj and Sampathila, Niranajana and Prabhu, Srikanth and Bhandage, Venkatesh and Hegde, Govardhan K},
  journal={Applied System Innovation},
  volume={6},
  number={2},
  pages={32},
  year={2023},
  publisher={MDPI}
}

@inproceedings{bell2022s,
  title={It’s just not that simple: an empirical study of the accuracy-explainability trade-off in machine learning for public policy},
  author={Bell, Andrew and Solano-Kamaiko, Ian and Nov, Oded and Stoyanovich, Julia},
  booktitle={Proceedings of the 2022 ACM conference on fairness, accountability, and transparency},
  pages={248--266},
  year={2022}
}

@article{alsinglawi2022explainable,
  title={An explainable machine learning framework for lung cancer hospital length of stay prediction},
  author={Alsinglawi, Belal and Alshari, Osama and Alorjani, Mohammed and Mubin, Omar and Alnajjar, Fady and Novoa, Mauricio and Darwish, Omar},
  journal={Scientific reports},
  volume={12},
  number={1},
  pages={607},
  year={2022},
  publisher={Nature Publishing Group UK London}
}

@article{petch2022opening,
  title={Opening the black box: the promise and limitations of explainable machine learning in cardiology},
  author={Petch, Jeremy and Di, Shuang and Nelson, Walter},
  journal={Canadian Journal of Cardiology},
  volume={38},
  number={2},
  pages={204--213},
  year={2022},
  publisher={Elsevier}
}

@article{mangalathu2022machine,
  title={Machine-learning interpretability techniques for seismic performance assessment of infrastructure systems},
  author={Mangalathu, Sujith and Karthikeyan, Karthika and Feng, De-Cheng and Jeon, Jong-Su},
  journal={Engineering Structures},
  volume={250},
  pages={112883},
  year={2022},
  publisher={Elsevier}
}

@article{xu2022review,
  title={Review on interpretable machine learning in smart grid},
  author={Xu, Chongchong and Liao, Zhicheng and Li, Chaojie and Zhou, Xiaojun and Xie, Renyou},
  journal={Energies},
  volume={15},
  number={12},
  pages={4427},
  year={2022},
  publisher={MDPI}
}

@inproceedings{hasib2022machine,
  title={A machine learning and explainable ai approach for predicting secondary school student performance},
  author={Hasib, Khan Md and Rahman, Farhana and Hasnat, Rashik and Alam, Md Golam Rabiul},
  booktitle={2022 IEEE 12th Annual Computing and Communication Workshop and Conference (CCWC)},
  pages={0399--0405},
  year={2022},
  organization={IEEE}
}

@inproceedings{nadeem2023sok,
  title={Sok: Explainable machine learning for computer security applications},
  author={Nadeem, Azqa and Vos, Dani{\"e}l and Cao, Clinton and Pajola, Luca and Dieck, Simon and Baumgartner, Robert and Verwer, Sicco},
  booktitle={2023 IEEE 8th European Symposium on Security and Privacy (EuroS\&P)},
  pages={221--240},
  year={2023},
  organization={IEEE}
}

@article{chen2024interpretable,
  title={Interpretable machine learning for imbalanced credit scoring datasets},
  author={Chen, Yujia and Calabrese, Raffaella and Martin-Barragan, Belen},
  journal={European Journal of Operational Research},
  volume={312},
  number={1},
  pages={357--372},
  year={2024},
  publisher={Elsevier}
}

@article{zhang2022interpretable,
  title={Interpretable machine learning models for crime prediction},
  author={Zhang, Xu and Liu, Lin and Lan, Minxuan and Song, Guangwen and Xiao, Luzi and Chen, Jianguo},
  journal={Computers, Environment and Urban Systems},
  volume={94},
  pages={101789},
  year={2022},
  publisher={Elsevier}
}

@article{hu2022interpretable,
  title={Interpretable machine learning for early prediction of prognosis in sepsis: a discovery and validation study},
  author={Hu, Chang and Li, Lu and Huang, Weipeng and Wu, Tong and Xu, Qiancheng and Liu, Juan and Hu, Bo},
  journal={Infectious diseases and therapy},
  volume={11},
  number={3},
  pages={1117--1132},
  year={2022},
  publisher={Springer}
}

@article{unvan2021market,
  title={Market basket analysis with association rules},
  author={{\"U}nvan, Y{\"u}ksel Akay},
  journal={Communications in Statistics-Theory and Methods},
  volume={50},
  number={7},
  pages={1615--1628},
  year={2021},
  publisher={Taylor \& Francis}
}

@article{pohlmeyer2022interpretable,
  title={Interpretable failure risk assessment for continuous production processes based on association rule mining},
  author={Pohlmeyer, Florian and Kins, Ruben and Cloppenburg, Frederik and Gries, Thomas},
  journal={Advances in Industrial and Manufacturing Engineering},
  volume={5},
  pages={100095},
  year={2022},
  publisher={Elsevier}
}

@article{troncoso2023new,
  title={A new approach based on association rules to add explainability to time series forecasting models},
  author={Troncoso-Garc{\'\i}a, AR and Mart{\'\i}nez-Ballesteros, Mar{\'\i}a and Mart{\'\i}nez-{\'A}lvarez, Francisco and Troncoso, Alicia},
  journal={Information Fusion},
  volume={94},
  pages={169--180},
  year={2023},
  publisher={Elsevier}
}

@article{bohmer2020mining,
  title={Mining association rules for anomaly detection in dynamic process runtime behavior and explaining the root cause to users},
  author={B{\"o}hmer, Kristof and Rinderle-Ma, Stefanie},
  journal={Information Systems},
  volume={90},
  pages={101438},
  year={2020},
  publisher={Elsevier}
}

@inproceedings{li2024interpretable,
  title={Interpretable Analysis of Production GPU Clusters Monitoring Data via Association Rule Mining},
  author={Li, Baolin and Samsi, Siddharth and Gadepally, Vijay and Tiwari, Devesh},
  booktitle={2024 IEEE International Parallel and Distributed Processing Symposium (IPDPS)},
  pages={337--349},
  year={2024},
  organization={IEEE}
}

@article{kusak2021apriori,
  title={Apriori association rule and K-means clustering algorithms for interpretation of pre-event landslide areas and landslide inventory mapping},
  author={Kusak, Lutfiye and Unel, Fatma Bunyan and Alptekin, Ayd{\i}n and Celik, Mehmet Ozgur and Yakar, Murat},
  journal={Open Geosciences},
  volume={13},
  number={1},
  pages={1226--1244},
  year={2021},
  publisher={De Gruyter}
}

@incollection{veerappa2023explaining,
  title={Explaining CNN classifier using association rule mining methods on time-series},
  author={Veerappa, Manjunatha and Anneken, Mathias and Burkart, Nadia and Huber, Marco F},
  booktitle={Explainable Deep Learning AI},
  pages={173--189},
  year={2023},
  publisher={Elsevier}
}

@article{yedjour2018symbolic,
  title={Symbolic interpretation of artificial neural networks based on multiobjective genetic algorithms and association rules mining},
  author={Yedjour, Dounia and Benyettou, Abdelkader},
  journal={Applied Soft Computing},
  volume={72},
  pages={177--188},
  year={2018},
  publisher={Elsevier}
}

@article{sankar2024study,
  title={A Study on the Explainability of Thyroid Cancer Prediction: SHAP Values and Association-Rule Based Feature Integration Framework.},
  author={Sankar, Sujithra and Sathyalakshmi, S},
  journal={Computers, Materials \& Continua},
  volume={79},
  number={2},
  pages={3111--3138},
  year={2024}
}

@inproceedings{parham2023explaining,
  title={Explaining the Behavior of Reinforcement Learning Agents Using Association Rules},
  author={Parham, Zahra and de Lille, Vi Tching and Cappart, Quentin},
  booktitle={International Conference on Learning and Intelligent Optimization},
  pages={107--120},
  year={2023},
  organization={Springer}
}

@inproceedings{agrawal1993mining,
  title={Mining association rules between sets of items in large databases},
  author={Agrawal, Rakesh and Imieli{\'n}ski, Tomasz and Swami, Arun},
  booktitle={Proceedings of the 1993 ACM SIGMOD international conference on Management of data},
  pages={207--216},
  year={1993}
}

@book{srikant1996fast,
  title={Fast algorithms for mining association rules and sequential patterns},
  author={Srikant, Ramakrishnan},
  year={1996},
  publisher={The University of Wisconsin-Madison}
}

@inproceedings{zaki1997new,
  title={New algorithms for fast discovery of association rules.},
  author={Zaki, Mohammed Javeed and Parthasarathy, Srinivasan and Ogihara, Mitsunori and Li, Wei and others},
  booktitle={KDD},
  volume={97},
  pages={283--286},
  year={1997}
}

@article{han2000mining,
  title={Mining frequent patterns without candidate generation},
  author={Han, Jiawei and Pei, Jian and Yin, Yiwen},
  journal={ACM sigmod record},
  volume={29},
  number={2},
  pages={1--12},
  year={2000},
  publisher={ACM New York, NY, USA}
}

@inproceedings{fortin1996object,
  title={An object-oriented approach to multi-level association rule mining},
  author={Fortin, Scott and Liu, Ling},
  booktitle={Proceedings of the fifth international conference on Information and knowledge management},
  pages={65--72},
  year={1996}
}

@article{wu2015review,
  title={A review on algorithms for maximum clique problems},
  author={Wu, Qinghua and Hao, Jin-Kao},
  journal={European Journal of Operational Research},
  volume={242},
  number={3},
  pages={693--709},
  year={2015},
  publisher={Elsevier}
}

@article{Karp1972,
  title={Reductibility among combinatorial problems},
  author={Karp, Richard M},
  journal={Complexity of computer computations},
  volume={1},
  number={1},
  pages={85--103},
  year={1972},
  publisher={Plenum Press New York}
}

@article{konc2022protein,
  title={Protein binding sites for drug design},
  author={Konc, Janez and Jane{\v{z}}i{\v{c}}, Du{\v{s}}anka},
  journal={Biophysical Reviews},
  volume={14},
  number={6},
  pages={1413--1421},
  year={2022},
  publisher={Springer}
}

@article{phillipson2023quantum,
  title={Quantum Computing in Telecommunication—A Survey},
  author={Phillipson, Frank},
  journal={Mathematics},
  volume={11},
  number={15},
  pages={3423},
  year={2023},
  publisher={MDPI}
}

@article{wang2022mini,
  title={A mini review of node centrality metrics in biological networks},
  author={Wang, Mengyuan and Wang, Haiying and Zheng, Huiru},
  journal={International Journal of Network Dynamics and Intelligence},
  volume={1},
  number={1},
  pages={99--110},
  year={2022}
}

@article{cavique2018data,
  title={A data reduction approach using hypergraphs to visualize communities and brokers in social networks},
  author={Cavique, Lu{\'\i}s and Marques, Nuno C and Gon{\c{c}}alves, Ant{\'o}nio},
  journal={Social Network Analysis and Mining},
  volume={8},
  pages={1--17},
  year={2018},
  publisher={Springer}
}

@inproceedings{azriel2019sok,
  title={Sok: An overview of algorithmic methods in IC reverse engineering},
  author={Azriel, Leonid and Ginosar, Ran and Mendelson, Avi},
  booktitle={Proceedings of the 3rd ACM Workshop on Attacks and Solutions in Hardware Security Workshop},
  pages={65--74},
  year={2019}
}

@article{li2017minimization,
  title={On minimization of the number of branches in branch-and-bound algorithms for the maximum clique problem},
  author={Li, Chu-Min and Jiang, Hua and Many{\`a}, Felip},
  journal={Computers \& Operations Research},
  volume={84},
  pages={1--15},
  year={2017},
  publisher={Elsevier}
}

@inproceedings{li2010efficient,
  title={An efficient branch-and-bound algorithm based on maxsat for the maximum clique problem},
  author={Li, Chu-Min and Quan, Zhe},
  booktitle={Proceedings of the AAAI Conference on Artificial Intelligence},
  volume={24},
  pages={128--133},
  year={2010}
}

@inproceedings{jiang2016combining,
  title={Combining efficient preprocessing and incremental MaxSAT reasoning for MaxClique in large graphs},
  author={Jiang, Hua and Li, Chu-Min and Many{\`a}, Felip},
  booktitle={Proceedings of the twenty-second European conference on artificial intelligence},
  pages={939--947},
  year={2016}
}

@article{san2019new,
  title={A new branch-and-bound algorithm for the maximum weighted clique problem},
  author={San Segundo, Pablo and Furini, Fabio and Artieda, Jorge},
  journal={Computers \& Operations Research},
  volume={110},
  pages={18--33},
  year={2019},
  publisher={Elsevier}
}

@article{wu2012multi,
  title={Multi-neighborhood tabu search for the maximum weight clique problem},
  author={Wu, Qinghua and Hao, Jin-Kao and Glover, Fred},
  journal={Annals of Operations Research},
  volume={196},
  pages={611--634},
  year={2012},
  publisher={Springer}
}

@inproceedings{wang2016two,
  title={Two efficient local search algorithms for maximum weight clique problem},
  author={Wang, Yiyuan and Cai, Shaowei and Yin, Minghao},
  booktitle={Proceedings of the AAAI Conference on Artificial Intelligence},
  volume={30},
  pages={805--811},
  year={2016}
}

@article{wang2020sccwalk,
  title={SCCWalk: An efficient local search algorithm and its improvements for maximum weight clique problem},
  author={Wang, Yiyuan and Cai, Shaowei and Chen, Jiejiang and Yin, Minghao},
  journal={Artificial Intelligence},
  volume={280},
  pages={103230},
  year={2020},
  publisher={Elsevier}
}

@article{angelini2021mismatching,
  title={Mismatching as a tool to enhance algorithmic performances of Monte Carlo methods for the planted clique model},
  author={Angelini, Maria Chiara and Fachin, Paolo and de Feo, Simone},
  journal={Journal of Statistical Mechanics: Theory and Experiment},
  volume={2021},
  number={11},
  pages={113406},
  year={2021},
  publisher={IOP Publishing}
}

@article{karalias2020erdos,
  title={Erdos goes neural: an unsupervised learning framework for combinatorial optimization on graphs},
  author={Karalias, Nikolaos and Loukas, Andreas},
  journal={Advances in Neural Information Processing Systems},
  volume={33},
  pages={6659--6672},
  year={2020}
}

@article{min2022can,
  title={Can hybrid geometric scattering networks help solve the maximum clique problem?},
  author={Min, Yimeng and Wenkel, Frederik and Perlmutter, Michael and Wolf, Guy},
  journal={Advances in Neural Information Processing Systems},
  volume={35},
  pages={22713--22724},
  year={2022}
}

@article{sanokowski2024variational,
  title={Variational annealing on graphs for combinatorial optimization},
  author={Sanokowski, Sebastian and Berghammer, Wilhelm and Hochreiter, Sepp and Lehner, Sebastian},
  journal={Advances in Neural Information Processing Systems},
  volume={36},
  year={2024}
}

@inproceedings{haverly2021implementation,
  title={Implementation of Grover’s algorithm to solve the maximum clique problem},
  author={Haverly, A and L{\'o}pez, S},
  booktitle={2021 IEEE Computer Society Annual Symposium on VLSI (ISVLSI)},
  pages={441--446},
  year={2021},
  organization={IEEE}
}

@article{li2024parameter,
  title={A parameter-independent algorithm of finding maximum clique with Seidel continuous-time quantum walks},
  author={Li, Xi and Chen, Xiao and Hu, Shouwei and Xu, Juan and Liu, Zhihao},
  journal={Iscience},
  volume={27},
  number={2},
  year={2024},
  publisher={Elsevier}
}

@inproceedings{pelofske2019solving,
  title={Solving large maximum clique problems on a quantum annealer},
  author={Pelofske, Elijah and Hahn, Georg and Djidjev, Hristo},
  booktitle={Quantum Technology and Optimization Problems: First International Workshop, QTOP 2019, Munich, Germany, March 18, 2019, Proceedings 1},
  pages={123--135},
  year={2019},
  organization={Springer}
}

@article{pardalos1994maximum,
  title={The maximum clique problem},
  author={Pardalos, Panos M and Xue, Jue},
  journal={Journal of global Optimization},
  volume={4},
  pages={301--328},
  year={1994},
  publisher={Springer}
}

@article{bomze1999maximum,
  title={The maximum clique problem},
  author={Bomze, Immanuel M and Budinich, Marco and Pardalos, Panos M and Pelillo, Marcello},
  journal={Handbook of Combinatorial Optimization: Supplement Volume A},
  pages={1--74},
  year={1999},
  publisher={Springer}
}

@article{feige2004approximating,
  title={Approximating maximum clique by removing subgraphs},
  author={Feige, Uriel},
  journal={SIAM Journal on Discrete Mathematics},
  volume={18},
  number={2},
  pages={219--225},
  year={2004},
  publisher={SIAM}
}

@article{engebretsen2003towards,
  title={Towards optimal lower bounds for clique and chromatic number},
  author={Engebretsen, Lars and Holmerin, Jonas},
  journal={Theoretical Computer Science},
  volume={299},
  number={1-3},
  pages={537--584},
  year={2003},
  publisher={Elsevier}
}

@inproceedings{zuckerman2006linear,
  title={Linear degree extractors and the inapproximability of max clique and chromatic number},
  author={Zuckerman, David},
  booktitle={Proceedings of the thirty-eighth annual ACM symposium on Theory of computing},
  pages={681--690},
  year={2006}
}

@article{chalermsook2020gap,
  title={From gap-exponential time hypothesis to fixed parameter tractable inapproximability: Clique, dominating set, and more},
  author={Chalermsook, Parinya and Cygan, Marek and Kortsarz, Guy and Laekhanukit, Bundit and Manurangsi, Pasin and Nanongkai, Danupon and Trevisan, Luca},
  journal={SIAM Journal on Computing},
  volume={49},
  number={4},
  pages={772--810},
  year={2020},
  publisher={SIAM}
}

@misc{gurobi,
  author = {{Gurobi Optimization, LLC}},
  title = {{Gurobi Optimizer Reference Manual}},
  year = 2023
}

@inproceedings{yanardag2015deep,
  title={Deep graph kernels},
  author={Yanardag, Pinar and Vishwanathan, SVN},
  booktitle={Proceedings of the 21th ACM SIGKDD international conference on knowledge discovery and data mining},
  pages={1365--1374},
  year={2015}
}

@inproceedings{yan2008mining,
  title={Mining significant graph patterns by leap search},
  author={Yan, Xifeng and Cheng, Hong and Han, Jiawei and Yu, Philip S},
  booktitle={Proceedings of the 2008 ACM SIGMOD international conference on Management of data},
  pages={433--444},
  year={2008}
}


\end{document}